\documentclass{article}

\usepackage{microtype}
\usepackage{graphicx}
\usepackage{subfigure}
\usepackage{booktabs} 
\usepackage{amsthm}

\usepackage{enumitem}
\setitemize{noitemsep,topsep=0pt,parsep=0pt,partopsep=0pt}

\newcommand{\Xopt}{{\hat{\mX}}^*}

\newtheorem{thm}{Theorem}

\usepackage[accepted]{icml2019}

\usepackage{amsmath,amssymb}
\usepackage{color,soul}

\usepackage{amsmath,amsfonts,bm}

\def\eqref#1{equation~\ref{#1}}

\def\1{\bm{1}}

\def\vw{{\bm{w}}}
\def\vx{{\bm{x}}}
\def\vy{{\bm{y}}}

\def\mA{{\bm{A}}}

\def\mD{{\bm{D}}}

\def\mF{{\bm{F}}}

\def\mH{{\bm{H}}}
\def\mI{{\bm{I}}}

\def\mP{{\bm{P}}}

\def\mX{{\bm{X}}}

\DeclareMathAlphabet{\mathsfit}{\encodingdefault}{\sfdefault}{m}{sl}
\SetMathAlphabet{\mathsfit}{bold}{\encodingdefault}{\sfdefault}{bx}{n}

\def\gG{{\mathcal{G}}}

\def\gO{{\mathcal{O}}}

\def\sR{{\mathbb{R}}}

\newcommand{\E}{\mathbb{E}}

\usepackage[textsize=footnotesize,disable]{todonotes}
\usepackage{url}

\newcommand{\ptitlenoskip}[1]{\noindent{\bf #1.}}

\newcommand{\ha}{\hat{a}}
\newcommand{\hmA}{\hat{\mA}}
\newcommand{\cN}{\mathcal{N}}
\newcommand{\cL}{\mathcal{L}}
\newcommand{\Popt}{{\hat{P}}^*}

\newcommand{\lij}{\ensuremath{\{i,j\}}}
\newcommand{\lik}{\ensuremath{\{i,k\}}}

\icmltitlerunning{ExplaiNE: An Approach for Explaining Network Embedding-based Link Predictions}

\begin{document}

\twocolumn[
\icmltitle{ExplaiNE: An Approach for Explaining\\Network Embedding-based Link Predictions}




\begin{icmlauthorlist}
\icmlauthor{Bo Kang}{ghent}
\icmlauthor{Jefrey Lijffijt}{ghent}
\icmlauthor{Tijl De Bie}{ghent}
\end{icmlauthorlist}

\icmlaffiliation{ghent}{Department of Electronics and Information Systems (ELIS){, IDLab}, Ghent University, Ghent, Belgium}

\icmlcorrespondingauthor{Bo Kang}{bo.kang@ugent.be}
\icmlcorrespondingauthor{Jefrey Lijffijt}{jefrey.lijffijt@ugent.be}
\icmlcorrespondingauthor{Tijl De Bie}{tijl.debie@UGent.be}

\icmlkeywords{Explainability, Network Embedding}

\vskip 0.3in
]



\printAffiliationsAndNotice{}  

\begin{abstract}
  Networks are powerful data structures, but are challenging to work with for conventional machine learning methods.
	Network Embedding (NE) methods attempt to resolve this
	by learning vector representations for the nodes,
	for subsequent use in downstream machine learning tasks.

	Link Prediction (LP) is one such downstream machine learning task that is an important use case and popular benchmark for NE methods.
	Unfortunately, while NE methods perform exceedingly well at this task,
	they are lacking in transparency as compared to simpler LP approaches.

	We introduce ExplaiNE, an approach to offer counterfactual explanations for NE-based LP methods,
	by identifying existing links in the network that explain the predicted links.
	ExplaiNE is applicable to a broad class of NE algorithms.
	An extensive empirical evaluation for the NE method `Conditional Network Embedding' in particular
	demonstrates its accuracy and scalability.
\end{abstract}

\section{Introduction\label{sec:introduction}}

Network embeddings (NEs) have exploded in popularity in both the machine learning and data mining communities. By mapping a network's nodes into a vector space, NEs enable the application of a variety of machine learning methods on networks for important tasks such as link prediction (LP): the task to predict whether nodes are likely to be(come) connected in incomplete or evolving networks. LP has wide-ranging applications, for friendship recommendations, recommender systems, knowledge graph completion, etc.
While there are numerous conventional LP methods that predict links based on heuristic statistics computed over networks (e.g., based on the number of common neighbors) \citep[see, e.g.,][]{martinez2017survey}, recently proposed NE-based methods typically outperform those heuristic approaches \citep[e.g.,][]{grover2016node2vec, kang2018conditional}.

While the superior performance of NE-based LP methods is an advantage, a major disadvantage is that they do not easily allow for human-intelligible explanations of the predicted links. Yet, the ability to understand link predictions is important and useful for several reasons: (a) recommender systems that provide explanations are more easily trusted and more effective, (b) it allows data analysts to have a better understanding of the network characteristics such as node features and network dynamics, (c) transparency of automated processing systems is required in a growing number of regulations, and explanations can increase transparency.

We present ExplaiNE
, a mathematically principled counterfactual reasoning approach for explaining NE-based link predictions. In its simplest form, ExplaiNE quantifies how the probability of a predicted link \lij{} would be affected by weakening an existing link \lik{}. Links \lik{} that after weakening most strongly reduce the probability of the predicted link \lij{} then serve as counterfactual explanations. 

\begin{figure}[t]
  \centering
  \includegraphics[width=0.5\textwidth]{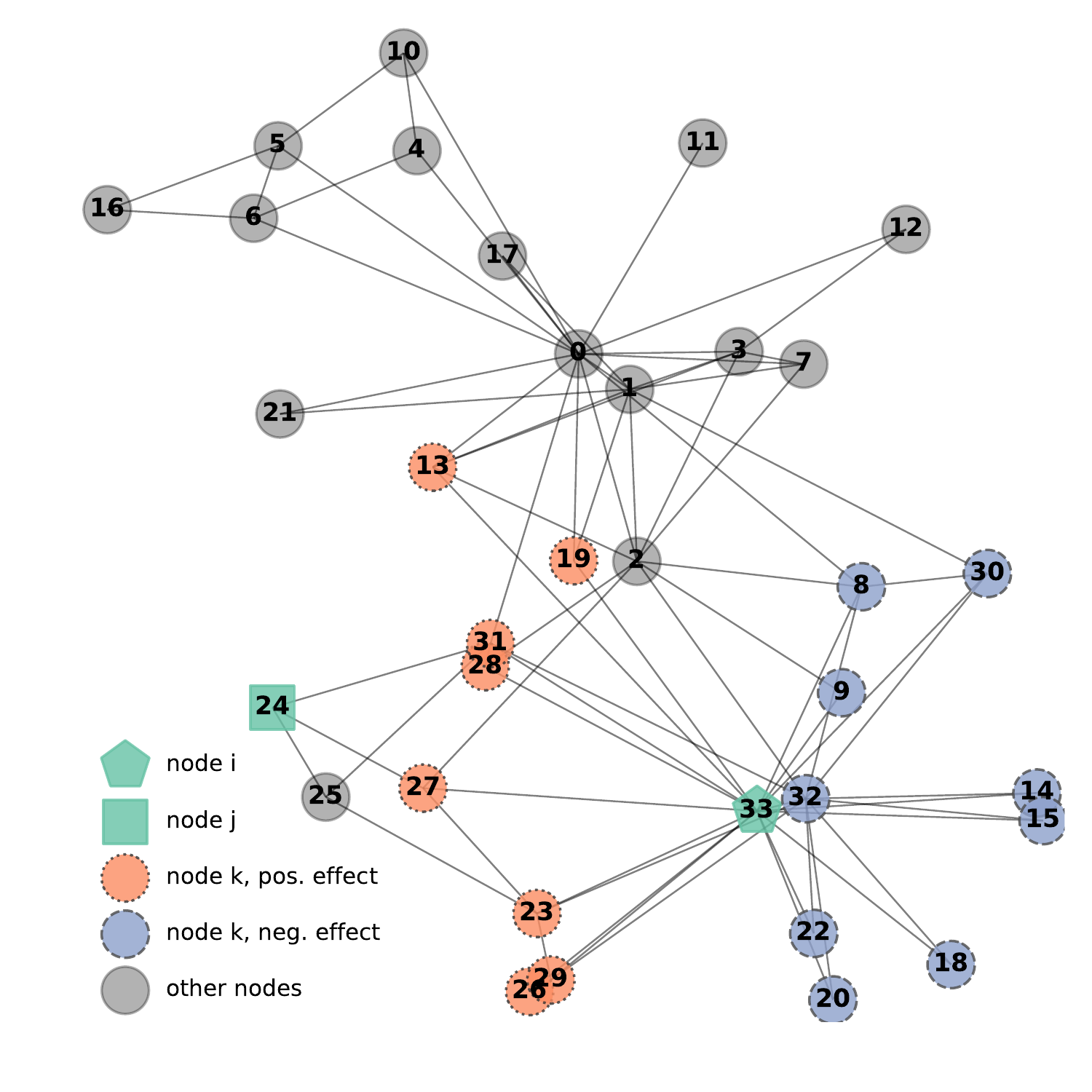}
	\vskip -0.7cm
\caption{In Zachary's karate club network, we explain the predicted link between $i=33$ and $j=24$. The colored nodes with dashed or dotted edges are the neighbors of node $33$. According to ExplaiNE, the links to the orange nodes with dotted edge have a positive effect on the probability of link $\{i,j\}$ to exist, while the effect of the links to blue nodes with dashed edge is negative. \label{fig:intro_illustr_karate}}
\end{figure}

\ptitlenoskip{Example} We show the idea of ExplaiNE on Zachary's karate club network \citep{zachary1977information}. The network consists of $34$ karate club members (nodes), with $78$ friendship links. An NE-based LP method\footnote{The embedding used is 2-dimensional and derived using Conditional Network Embedding \citep{kang2018conditional}.} to predict a link for node $i=33$ (green pentagon) indicates a high probability link to node $j=24$ (green square). Figure~\ref{fig:intro_illustr_karate} visualizes the embedding and highlights which existing links incident to $i$ ExplaiNE deems explanatory for this prediction in a positive (orange circle, dotted edge) or negative sense (blue circle, dashed edge). It concludes this because weakening links to the orange nodes would reduce the link probability \lij{}, whereas weakening links to the blue nodes would increase it. Note that these effects are quite intuitive given the geometry of the embedding: the orange nodes `pull' node $33$ closer to $24$
, while the blue nodes pull node $33$ away from $24$.

ExplaiNE is first derived as generically as possible, allowing for explanations not only in terms of links incident to the predicted link, but also in terms of other links as well as non-links.
We then reduce its scope to explanations of the type used in the example above (i.e., only incident links), and make an approximation (which we justify empirically), in order to obtain a still generic but highly scalable approach.

Next we apply ExplaiNE to Conditional Network Embedding \citep[CNE;][]{kang2018conditional}, a recent state-of-the-art NE method.
The application of ExplaiNE to CNE is particularly transparent, thanks to the mathematical elegance of the CNE model and its straightforward use in LP, requiring no training once the embedding is found.
We also outline how ExplaiNE can be applied to NE methods based on skip gram with negative sampling-based such as LINE \citep{tang2015line}, DeepWalk \citep{perozzi2014deepwalk}, PTE \citep{tang2015pte}, and node2vec \citep{grover2016node2vec}.

\ptitlenoskip{Contributions} The main contributions are:
\begin{itemize}
  \item
    ExplaiNE, a mathematically principled counterfactual reasoning approach for explaining link predictions based on network embeddings (Sec.~\ref{sec:method_explaine}).
  \item
    A scalable tight approximation of ExplaiNE (Sec.~\ref{sec:appr}).
  \item
    A detailed application of ExplaiNE to CNE (Sec.~\ref{sec:explaine_cne})
	\item
	  An outline of how to apply ExplaiNE to NE methods based on skip gram with negative sampling (Sec.~\ref{sec:explaine_other})
  \item
    Quantitative and run time analyses showing the stability and scalability of the approximation. (Sec.~\ref{sec:exp_approx},\ref{sec:exp_runtime})
  \item
    Qualitative and quantitative realistic case studies confirming the usefulness of ExplaiNE. (Sec.~\ref{sec:exp_qualitative},\ref{sec:exp_quantitative})
\end{itemize}


\section{Methods\label{sec:method}}

To introduce ExplaiNE in full generality,
we first provide a simple but generic description of NE-based link prediction methods in Section~\ref{sec:method_ne_lp}.
We then formalize ExplaiNE in a generic manner in Section~\ref{sec:method_explaine}
, before describing a scalable approximation in Section~\ref{sec:appr}. In Section~\ref{sec:explaine_cne} we develop ExplaiNE in detail for CNE. In Section \ref{sec:explaine_other} we outline how ExplaiNE can also be applied to other popular NE methods
.
But before all that, we first introduce some notation.

An undirected network is denoted $\gG = (V, E)$ where $V$ is a set of $n=|V|$ nodes and $E\subseteq \binom{V}{2}$ is the set of links (also known as edges). A link is denoted by an unordered node pair $\{i,j\} \in E$. Let $\mA$ denote the adjacency matrix, with element $a_{ij} = 1$ for $\{i,j\} \in E$ and $a_{ij} = 0$  otherwise. The symbol $\hmA$ will be used to denote the adjacency matrix of a particular observed network. NE methods find a mapping $f: V \to \sR^d$ from nodes to $d$-dimensional real vectors. An embedding is denoted as $\mX = (\vx_1, \vx_2,\ldots,\vx_n)' \in \sR^{n \times d}$, with $\mX^*$ denoting an optimal embedding for adjacency matrix $\mA$ (suppressing the dependency of $\mX$ on $\mA$ for conciseness---see below), and similarly $\Xopt$ optimal for $\hmA$. 

\subsection{Network Embedding-based Link Predictions\label{sec:method_ne_lp}}

All well-known NE methods aim to find an embedding $\mX^*$ for given graph $\gG$ (with adjacency matrix $\mA$) that maximizes a continuously differentiable\footnote{Note that, although NE methods are often described for unweighted networks (i.e., a binary adjacency matrix),
the objective $\cL(\mA, \mX)$ is often continuously differentiable also w.r.t. the adjacency matrix $\mA$.
This is required for ExplaiNE to be applicable, but as we will see this requirement is often satisfied.} objective function $\cL(\mA, \mX)$ for the given adjacency matrix $\mA$.
Thus $\mX^*$ must satisfy the following necessary condition for optimality:
\begin{align}\label{eq:opt_condt}
  \nabla_{\mX} \cL(\mA, \mX^*) = \mathbf{0}.
\end{align}
Defining $\mF(\mA,\mX)\triangleq\nabla_{\mX}\cL(\mA, \mX)$,
the optimal embedding $\mX^*$ is thus a solution to $\mF\left(\mA, \mX^*\right)=\mathbf{0}$.

Based on an embedding $\mX$, it is common to predict the existence of a link between any pair of nodes $i$ and $j$ by computing a link probability (or other score) $g_{ij}(\mX)$, using a differentiable function $g_{ij}:\sR^{nd}\rightarrow\sR$.
In practice, $g_{ij}$ often only depends on the embeddings $\vx_i$ and $\vx_j$ of $i$ and $j$, and often it can be written as $g_{ij}(\mX)=g(\vx_i,\vx_j)$ for some function $g:\sR^d\times\sR^d\rightarrow\sR$.
It is often found by training a classifier
(e.g., logistic regression) on a set of known linked and unlinked node pairs (see Sec.~\ref{sec:explaine_other}),
but sometimes it follows directly from the NE model (e.g., for CNE).

We also introduce the function $g^*_{ij}:\sR^{n\times n}\rightarrow\sR$ defined as $g^*_{ij}(\mA)\triangleq g_{ij}(\mX^*)$
where $\mX^*$ is optimal w.r.t. $\mA$.
I.e., $g^*_{ij}$ directly computes the link probability w.r.t. an optimal embedding for a specified adjacency matrix.

\subsection{ExplaiNE as a generic approach\label{sec:method_explaine}}

ExplaiNE uses a counterfactual reasoning approach to explain link predictions based on a NE. Namely, it quantifies the change of the link probability (or other score) of a node pair $\{i,j\}$ if the presence of a link between a given pair of nodes $\{k,l\}$ were to be altered.
Consider first the situation where $\{k,l\}\in E$. Then, if removing the link between them strongly decreases the probability of a link between $i$ and $j$, the link $\{k,l\}$ is a good counterfactual explanation of this predicted link.
Conversely, consider the situation where $\{k,l\}\not\in E$. Then, if adding a link between them strongly decreases the probability of a link between $i$ and $j$, it is the absence of a link between $k$ and $l$ that is a good counterfactual explanation of this predicted link.

Intuitively, adding or removing an existing link will alter the probability of a link between $i$ and $j$ because it will alter the optimal embedding, which in turn will change the link probability of the target pair.
For the ExplaiNE strategy to be effective, we must be able to compute and combine these two effects in an efficient manner.

A naive approach would be to recompute the embedding with a link added or removed,
and to quantify how much this changes the probability of a link between $i$ and $j$.
However, recomputing the embedding is computationally demanding,
and is practically impossible to do even for a moderate number of pairs $\{k,l\}$.
Moreover, even adding or removing a single link can dramatically change the optimization landscape.
As there are potentially many local optima,
this can change the optimal embedding entirely (even if initialized with the original embedding),
making a change in link probability erratic and hard to interpret.

Instead, ExplaiNE investigates the effect of an \emph{infinitesimal} change to $a_{kl}$ around its observed value $\ha_{kl}$,
on the link probability as computed by $g^*_{ij}$. 
Specifically, ExplaiNE seeks explanations as node-pairs $\{k,l\}$ ($k\neq l$ and $\{k, l\} \neq \{i, j\}$) for which $\frac{\partial g_{ij}^*}{\partial a_{kl}}(\hmA)$ is large in absolute value, with a positive sign if $\ha_{kl}=1$ (as then decreasing $a_{kl}$ down from $\ha_{kl}=1$ by a small amount would maximally decrease $g_{ij}^*$), and with a negative sign if $\ha_{kl}=0$ (as then increasing $a_{kl}$ up from $\ha_{kl}=0$ by a small amount would maximally decrease $g_{ij}^*$). This can be done analytically. Indeed, applying the chain rule:
\begin{align}\label{eq:chainrule}
  \frac{\partial g_{ij}^*}{\partial a_{kl}}(\hmA) = \nabla_{{\mX}}g_{ij}\left(\Xopt\right)^T\cdot\frac{\partial \mX^*}{\partial a_{kl}}(\hmA).
\end{align}
For many NE methods the first factor can be computed analytically from the expression for $g_{ij}$, as we will see in the next subsections.
The second factor can be computed using the \emph{implicit function theorem} \citep[see, e.g.,][]{chiang1984fundamental}. Rephrased for our specific setting, this theorem states (note that we are overloading the symbol $\mX^*$ here to also signify a function):
\begin{thm}[Implicit function theorem]
Let $\mF:\sR^{n\times n}\times\sR^{n\times d}\rightarrow \sR^{n\times d}$ be a continuously differentiable function with arguments denoted $\mA\in\sR^{n\times n}$ and $\mX\in\sR^{n\times d}$. Moreover, let $\hmA$ and $\Xopt$ be such that $\mF(\hmA,\Xopt)=\mathbf{0}$. If the Jacobian matrix $\nabla_{\mX}\mF(\hmA, \Xopt)$ is invertible, then there exists an open set $S\subset\sR^{n\times n}$ with $\hmA\in S$ such that there exists a continuously differentiable function $\mX^*:S\rightarrow \sR^{n\times d}$ with:
\begin{align*}
\mX^*(\hmA)&=\Xopt,\ \mbox{and}\\
\mF(\mA,\mX^*(\mA))&=\mathbf{0}\ \mbox{for all}\ \mA\in S,
\end{align*}
and:
\begin{align*}
  \frac{\partial \mX^*}{\partial a_{kl}}(\mA) &= -(\nabla_{\mX}\mF(\mA, \mX^*(\mA)))^{-1}\cdot\frac{\partial\mF}{\partial a_{kl}}(\mA, \mX^*(\mA)).
\end{align*}
\end{thm}
It is the latter expression, evaluated at $\hmA$, that we need in in order to evaluate Eq.~(\ref{eq:chainrule}).
Note that the Jacobian $\nabla_{\mX}\mF$ is in fact the Hessian of $\cL$ with respect to $\mX$.
This means that $\nabla_{\mX}\mF(\hmA,\Xopt)$ is negative definite (as $\Xopt$ is optimal for $\hmA$).
While for some NE-methods it may not be \emph{strictly} negative definite and thus not invertible as required by the theorem (because, e.g., any translation of $\Xopt$ may be equally optimal according to $\cL$), this situation can be avoided by adding a regularizer to $\cL$ on, e.g., the Frobenius norm of $\Xopt$ with very small weight. Without going into detail, we note that as this regularization constant approaches zero, this becomes equivalent with using the pseudo-inverse of the Hessian, instead of its inverse. This is the approach we have taken whenever this situation arose.
Denoting this Hessian evaluated at $\hmA$ and $\Xopt$ as $\mH$, can thus write:
\begin{align}\label{eq:ift}
  \frac{\partial \mX^*}{\partial a_{kl}}(\hmA) &=-\mH^{-1}\cdot\frac{\partial\mF}{\partial a_{kl}}(\hmA, \Xopt).
\end{align}

Putting Eqs.~(\ref{eq:chainrule}) and (\ref{eq:ift}) together, we now can compute the derivative of $g^*_{ij}$ with respect to $a_{kl}$ as follows:
\begin{align}\label{eq:exact_der}
\frac{\partial g_{ij}^*}{\partial a_{kl}}(\hmA) &=-\nabla_{{\mX}}g_{ij}\left(\Xopt\right)^T\cdot\mH^{-1}\cdot\frac{\partial\mF}{\partial a_{kl}}(\hmA, \Xopt).
\end{align}
For efficiency, one can compute the partial derivatives for a given predicted link $\{i,j\}$ and for all pairs $\{k,l\}$ by pre-computing the vector $\nabla_{{\mX}}g_{ij}\left(\Xopt\right)^T\cdot\mH^{-1}$ by solving a linear system with $nd$ variables and equations, and right multiplying it with the vectors $\frac{\partial\mF}{\partial a_{kl}}(\hmA, \Xopt)$ which depend on $k$ and $l$.
Unfortunately, the computational cost of solving this linear system is $\gO((nd)^3)$ in practice, limiting scalability both in network size and dimensionality.
Thus, while this is a clear improvement over the naive approach,
it is still not sufficient for realistic network sizes.
The next subsection describes how to make ExplaiNE tractable also for large networks and dimensionalities.

\subsection{Making ExplaiNE scalable}\label{sec:appr}

First, we choose to focus on explanations in terms of linked pairs $\{k,l\}$,
rather than in terms of unlinked pairs.
Such positive explanations are arguably more insightful than negative ones,
and especially in sparse networks.
Second, experiments (see supplement Sec.~4.1)\todo[inline]{We really should have investigated the following emperically.} show that the best explanation for a predicted link $\{i, j\}$ for a node $i$, tends to be a link $\{k, l\}$ that is incident to node $i$, i.e., for which $l=i$. This is arguably because links adjacent to node $i$ affect the link probability $g_{ij}^*(\hmA)$ by directly affecting the embedding $\vx^*_i$, whereas links not incident to $i$ are likely to have a secondary effect only. Besides this, we also believe that nodes incident to $i$ are likely to be more meaningful from node $i$'s perspective than other links, in practical applications.
Thus, we can restrict ourselves to seeking an explanation for a predicted link from node $i$ to node $j$
in terms of an existing link $\{i,k\}$ for which $\frac{\partial g_{ij}^*}{\partial\ha_{ik}}(\hmA)$ is large and positive.

Third, we consider only NE methods where $g_{ij}(\mX^*)$ only depends on $\vx_i^*$ and $\vx_j^*$.\footnote{This is true for all NE methods we are aware of, and thus hardly a limitation at all.} Thus, Eq.~(\ref{eq:chainrule}) can be written as:
\begin{align}\label{eq:chainrule2}
  \frac{\partial g_{ij}^*}{\partial a_{ik}}(\hmA) &= \nabla_{{\vx_i}}g_{ij}\left(\Xopt\right)^T\cdot\frac{\partial \vx^*_i}{\partial a_{ik}}(\hmA) \\
	&+ \nabla_{{\vx_j}}g_{ij}\left(\Xopt\right)^T\cdot\frac{\partial \vx^*_j}{\partial a_{ik}}(\hmA).\nonumber
\end{align}

Finally, we make an approximation
inspired by the fact that changing $a_{ik}$ will have a direct effect on the optimal embeddings $\vx^*_i$ and $\vx^*_k$, but only indirectly (and thus typically less so) on the embedding of the other nodes---including on $\vx^*_j$.
This means that the second term in Eq.~(\ref{eq:chainrule2}) can be neglected.

What remains to be computed is thus $\frac{\partial \vx^*_i}{\partial a_{ik}}(\hmA)$.
To do so, we consider the optimality condition of the embedding w.r.t. $\vx_i^*$ alone,
considering all other node embeddings fixed to their optimal embeddings in $\Xopt$ for the observed $\hmA$.
Letting $\Xopt_{(i)}$ denote the set of $\hat\vx_l$ with $l\neq i$
, this optimality condition can be written as:
\begin{align*}
\nabla_{\vx_i}\cL(\hmA,\vx_i,\Xopt_{(i)})=\mathbf{0}.
\end{align*}

For conciseness, let us define $\hat\mF_i(\mA, \vx_i)\triangleq \nabla_{\vx_i}\cL(\mA,\vx_i,\Xopt_{(i)})$. Optimality of $\hat\vx_i^*$ given the observed network $\hmA$ then requires that $\hat\mF_i(\hmA, \hat\vx_i^*)=\mathbf{0}$. We can now use the implicit function theorem on this optimality condition to approximate $\frac{\partial \vx^*_i}{\partial a_{ik}}$ as:
\begin{align}\label{eq:ift2}
\frac{\partial \vx^*_i}{\partial a_{ik}}(\hat{a}_{ik}) &= -\mH_i^{-1}\cdot\frac{\partial \hat\mF_i}{\partial a_{ik}}.
\end{align}
Here, $\mH_i = \nabla_{\vx_i} \hat\mF_i(\hmA, \hat\vx_i^*)$ is the Jacobian of $\hat\mF_i$ or equivalently the Hessian of $\cL$ w.r.t. $\vx_i$, evaluated at $(\hmA,\Xopt)$.

Putting Eqs.~(\ref{eq:ift2}) and (\ref{eq:chainrule2}) (neglecting the second term as discussed) together,
this yields:
\begin{align}\label{eq:approx_der}
\frac{\partial g_{ij}^*}{\partial a_{ik}}(\hmA) &= -\nabla_{{\vx_i}}g_{ij}\left(\Xopt\right)^T\cdot
\mH_i^{-1} \cdot\frac{\partial \hat\mF_i}{\partial a_{ik}}(\hmA,\hat\vx_i^*).
\end{align}
%
Comparing Eq.~(\ref{eq:exact_der}) with Eq.~(\ref{eq:approx_der}) reveals the dramatic complexity reduction achieved:
Inverting $\mH_i\in\sR^{d\times d}$ has a practical complexity of only $\gO(d^3)$, which is entirely feasible given common dimensionalities used in the literature (often 128). 
The experiments will validate that the approximations made are entirely justified in practice.

\subsection{ExplaiNE for Conditional Network Embedding\label{sec:explaine_cne}}

We now apply the generic ExplaiNE approach to Conditional Network Embedding (CNE), a specific NE method. Detailed derivations are deferred to the supplement Sec.~1.

CNE proposes a probability distribution for the network conditional on the embedding,
and finds the optimal embedding by maximum likelihood estimation.
Specifically, the objective function $\cL$ in CNE is the log-probability of the network conditioned on the embedding:
\begin{align*}
\cL(\hmA,\mX)=\log(P(\hmA|\mX)) &= \sum_{\{i,j\}:\ha_{ij}=1} \log{P_{ij}(a_{ij}=1|\mX)} \\
&+ \sum_{\{i,j\}:\ha_{ij}=0} \log{P_{ij}(a_{ij}=0|\mX)}.
\end{align*}
Here, the link probabilities $P_{ij}$ conditioned on the embedding are defined as follows:
\begin{align}\label{eq:posterior}
& P_{ij}(a_{ij}=1|\mX) = 1-P_{ij}(a_{ij}=0|\mX)=\\
& \frac{P_{\hmA,ij}\cN_{+,\sigma_1}(\|\vx_i-\vx_j\|)}{P_{\hmA,ij}\cN_{+,\sigma_1}(\|\vx_i-\vx_j\|) + (1-P_{\hmA,ij})\cN_{+,\sigma_2}(\|\vx_i-\vx_j\|)},\nonumber
\end{align}
where $\cN_{+,\sigma}$ denotes a half-Normal distribution \citep{leone1961folded} with spread parameter $\sigma$, $\sigma_2>\sigma_1=1$, and where $P_{\hmA,ij}$ is a prior probability for a link to exist between nodes $i$ and $j$ as inferred from the degrees of the nodes (or based on other information about the structure of the network)---see, e.g., \citet{adriaens2017subjectively, van2016subjective}.

CNE, being based on a probabilistic model for the graph conditioned on the embedding, naturally allows for LP using the probabilities $P_{ij}(\ha_{ij}=1|\mX)$. In other words, $g_{ij}(\mX)=P_{ij}(a_{ij}=1|\mX)$ as shown in Eq.~(\ref{eq:posterior}). Note that it depends on $\vx_i$ and $\vx_j$ alone, as required for the approximate version of ExplaiNE to be applicable (third assumption).

Next we show how to apply approximated ExplaiNE to CNE.\footnote{Due to the limited space, here we only show how to apply approximated ExplaiNE to CNE, as the exact version is not used in the experiments except for the experiment validating the approximated version. For the application of exact ExplainNE to CNE, we refer the reader to the supplement Sec.~2. From now on, we drop the modifier `approximated' when the context is clear.} First, we derive the optimality condition:
\begin{align*}
\hat\mF_i(\hmA, \hat\vx_i^*) &= \nabla_{\vx_i^*}\log(P(\hmA|\Xopt)) \\
&= \gamma\sum_{j\neq i} (\hat\vx_i^*-\hat\vx_j^*)\left(P\left(a_{ij}=1|\Xopt\right)-\ha_{ij}\right) \\
&= \mathbf{0}.
\end{align*}
Denoting $\gamma = \frac{1}{\sigma_1^2} - \frac{1}{\sigma_2^2}$, 
and $\hat{P}^*_{ij}\triangleq g_{ij}^*(\hmA)=P_{ij}(a_{ij} = 1 | \Xopt)$ (the probability of a link between $i$ and $j$ given the optimal embedding $\Xopt$ for $\hmA$),
we can now derive the three factors in Eq.~(\ref{eq:approx_der}):\footnote{Detailed derivations are provided in the supplementary material Sec.~1 due to space constraints.}
\begin{align*}
\nabla_{\vx_i}g_{ij}\left(\Xopt\right) &= -\gamma(\vx_i^*-\vx_j^*)\hat{P}^*_{ij}(1-\hat{P}^*_{ij}).\\
\mH_i &= \nabla_{\vx_i}\hat\mF_i(\hmA, \hat\vx_i^*)\\
&= \gamma
\mI\sum_{l\neq i} \left(P_{il}^*-\ha_{il}\right) \\
& -\gamma^2\sum_{l\neq i} (\vx_i^*-\vx_l^*)(\vx_i^*-\vx_l^*)'\hat{P}^*_{il}(1-\hat{P}^*_{il}).\\
\frac{\partial \hat\mF_i}{\partial a_{ik}}(\hmA,\hat\vx_i^*) &= \gamma(\vx_k^* - \vx_i^*).
\end{align*}

%
This means:
\begin{align*}
  \frac{\partial g_{ij}^*}{\partial a_{ik}}(\hmA)
	&= (\vx_i^*-\vx_j^*)^T\left(\frac{-\mH_i}{\gamma^2\hat{P}^*_{ij}(1-\hat{P}^*_{ij})}\right)^{-1}(\vx_i^* - \vx_k^*).
\end{align*}
Note that the Hessian should be invertible and negative definite, if $\hat\vx_i^*$ is indeed a local maximum.
Interestingly, this expression has an intuitive interpretation:
without the inverted Hessian, it would be an inner product between the distance of $\vx_i^*$ to the embeddings of both nodes $\vx_j^*$ and $\vx_k^*$,
indicating that the best explanation is located as far as possible in the direction of $\vx_j^*$ as seen from $\vx_i^*$. Yet, the Hessian modulates the metric and reduces the explanatory power in directions where there are lots of embedded nodes $l$ for which $\hat{P}^*_{il}(1-\hat{P}^*_{il})$ is large, i.e., for which the model is undecided whether there should be a link.

\todo[inline]{The math is overflowing in many equations, this will need to be fixed.}

\subsection{ExplaiNE for other NE methods\label{sec:explaine_other}}

Here we illustrate the generic applicability of ExplaiNE by outlining the steps of applying it to NE methods based on skip gram with negative sampling (SGNS) (e.g., LINE, PTE, DeepWalk, node2vec). In Sec.~3 of the supplement, we derive a concrete example for LINE \citep{tang2015line}.

In those methods, $g_{i,j}(\mX)=g(\vx_i,\vx_j)$, where $g\triangleq \sigma\circ h$ with $\sigma:\sR^d\rightarrow \sR$ a linear classifier (often logistic regression) applied to edge embeddings, whereby the embedding $h(\vx_i,\vx_j)$ of an edge $\{i,j\}$ is computed by applying an edge embedding operator $h: \sR^d\times \sR^d \to \sR^d$ (e.g., element-wise product) to the embeddings of the nodes at its end-points.

\citet{levy2014neural} and \citet{qiu2018network} found that SGNS-based NE methods all share the same objective:
\begin{align*}
\cL = \sum_{i=1}^{|V|}\sum_{j=1}^{|V|}\log\sigma(\vx_i\cdot\vy_j) + b\sum_{i=1}^{|V|}\E_{j'\sim P_N}\left[\log \sigma(-\vx_i\cdot\vy_{j'})\right],
\end{align*}
where $\vx_i$ is the target embedding of node $i$, $\vy_i$ is the embedding of node $j$ as context (usually discarded, node2vec does not differentiate target and context), $\sigma(\cdot)$ is a sigmoid function, $P_N$ is known as the noise that generates negative samples, and $b$ is the number of negative samples. Moreover, 
\citet{qiu2018network} showed that $\cL$ often has a closed form representation (or converges to one in probability).\todo[inline]{Bo, is this correct?}
This makes it possible to obtain an analytical expression of the NE optimality condition, and thus of the function $\mF(\mA,\mX)$.
Given this, both exact and approximated ExplaiNE can be derived.

\section{Experiments\label{sec:experiments}}

We investigated the following questions: \textbf{Q1} How does the approximation compare to the exact version? \textbf{Q2} Does ExplaiNE give sensible explanations? \textbf{Q3} Does the proposed method scale?

All experiments are based on CNE with parameters $\sigma_1=1$, $\sigma_2=2$. Any weights associated to the links in the networks are ignored. We used the following networks.

\ptitlenoskip{Game of Thrones' (GoT) network}\footnote{\url{https://github.com/mathbeveridge/asoiaf}} Consisting of $796$ characters (nodes) and $2823$ links between characters that are mentioned within $15$ words of one another in books 1-5. We used a $2$-dimensional embedding of this network to assess the quality of the approximated ExplaiNE approach.

\ptitlenoskip{DBLP co-authorship network \citep{tang2008arnetminer}}\footnote{DBLP dataset V10: \url{https://aminer.org/citation}} Containing papers published up to year 2017, from which we selected all papers published at ICML, NeurIPS, ICLR, JMLR, MLJ, KDD, ECML-PKDD, and DMKD. This results in 23,359 authors (nodes) and 20,545 papers, converted into 66,597 links between authors who co-authored at least one paper. We conducted both qualitative and quantitative evaluations on a $32$-dimensional embedding of this network.

\ptitlenoskip{MovieLens dataset \citep{harper2016movielens}}\footnote{\url{https://grouplens.org/datasets/movielens/100k/}} Containing 100,000 ratings by 943 users on 1,682 movies. The network is thus bipartite and consists of 943+1,682 nodes and 100,000 edges. The dataset also contains metadata such as title and genre, which we have used as external validation sources. We conducted qualitative and quantitative experiments on a $16$-dimensional embedding of this network.

In Sec.~\ref{sec:exp_approx} we analyze the quality of the approximation. In Sec.~\ref{sec:exp_qualitative}, we conduct a qualitative analysis of explanations on the DBLP and MovieLens networks. In Sec.~\ref{sec:exp_quantitative} we quantitatively analyze the quality of the explanations. Finally, in Sec.~\ref{sec:exp_runtime} we consider the scalability of ExplaiNE.

\subsection{Quality of the ExplaiNE approximation\label{sec:exp_approx}}

Before applying approximated ExplaiNE to real world dataset, we first evaluate the quality of the approximation (\textbf{Q1}). We will assess the extent to which the top $K$ explanations for a predicted link $\{i,j\}$ incident to a given node $i$, as given by approximated ExplaiNE, overlap with the top-$K$ explanations given by exact ExplaiNE.
Relevant parameters here are (1) the value of $K$ and (2) the number of neighbors.
As we consider only links to neighbors as candidate explanations, $K$ must be smaller than the number of neighbors of $i$. Moreover, if the number of neighbors is not much larger than $K$, a substantial overlap in the top-K explanations of the exact and approximate method is not surprising. Indeed, if $i$ has $m$ neighbors, two random subset of $K$ neighbors would share $l$ elements with probability $\binom{K}{l}\binom{m-K}{K-l}/\binom{m}{K}$, which is large for large $l$ if $m$ is not much larger than $K$.

Thus, we performed a stratified analysis, computing the size of the overlap of the top-$K$ explanations, aggregated in a histogram over nodes with a specific degree.
We did this on the GoT dataset for $K$ from $1$ to $5$
This experiment revealed that the top-1 is always identical between the approximated and exact versions, while the elements further in the ranked list very rarely swapped positions ($2$ to $3$ differences out of $796$ on ranks $2$,$3$,$4$, and $7$ differences out of $796$ for rank $5$, see supplement Sec.~4.2). 

In the supplement Sec.~4.3 we also compared the complete ranking of the neighbors between the approximated and exact ExplaiNE versions, and this for the most probable link for every node (i.e., seeking explanations for links that are actually present in the network). We computed the normalized Kendall tau distance\footnote{\url{https://en.wikipedia.org/wiki/Kendall_tau_distance}} between the ranked explanations given by approximated and exact ExplaiNE. 
The average normalized Kendall tau distance is $0.05\pm 0.08$. For comparison, the average Kendall tau distance between a random ranking and exact ExplaiNE is $0.51 \pm 0.15$.

Now confident in its accuracy, we can now evaluate the behavior of approximated ExplaiNE on two realistic networks.

\subsection{Qualitative evaluation\label{sec:exp_qualitative}}

Here we apply ExplaiNE to explain the predicted links in two real world networks (the DBLP co-authorship and the MovieLens rating networks) to assess whether ExplaiNE gives sensible explanations to the predicted links (\textbf{Q2}).

\ptitlenoskip{DBLP network} In the co-authorship network, a predicted link between authors $i$ and $j$ suggests a collaboration between them. While ExplaiNE uses no external information to provide its explanations for such suggested collaborations, our experiments indicate that such explanations tend to be existing collaborators working on a topic on which the suggested collaborator is active as well.
As an example, we predict links for ICML'19 general chair Eric P. Xing (node $i$), and compute the explanations for his top recommendation (node $j$): Adams Wei Yu. It turns out that the existing co-authors of Eric P. Xing identified by ExplaiNE as top-$5$ explanations for this recommendation (see Table~\ref{tb:dblp_explanation}) are either colleagues or co-authors of Adams Wei Yu, with a shared interest in large scale optimization and deep learning.

\begin{table}
  \caption{Predicted/recommended collaborations for Eric P. Xing. The top link (author: Adams Wei Yu) predicted by CNE are explained through co-authors of Eric P. Xing that are also colleagues or co-authors of Adams Wei Yu. The most relevant five co-authors of Eric P. Xing also cover major parts of Adams Wei Yu's research interests: large scale optimization and deep learning.}
  \label{tb:dblp_explanation}
\centering  \begin{tabular}{ c  l l}
    \multicolumn{1}{c}{Rank} & \multicolumn{1}{c}{Recommendations} &\multicolumn{1}{c}{\begin{tabular}{@{}c@{}}Explain: \\  `Adams Wei Yu' \end{tabular}}\\\hline
    1 & Adams Wei Yu & \multicolumn{1}{c}{Hao Su} \\\cline{1-2}
    2 & Jure Leskovec & \multicolumn{1}{|c}{Li Fei-Fei} \\
    3 & Sunita Sarawagi & \multicolumn{1}{|c}{Suvrit Sra} \\
    4 & Tong Zhang & \multicolumn{1}{|c}{Fan Li} \\
    5 & Soumen Chakrabarti & \multicolumn{1}{|c}{Wei Dai} \\
  \end{tabular}
\end{table}

\ptitlenoskip{MovieLens network} In the rating network, a predicted link between a user $i$ and movie $j$ amounts to a recommendation of movie $j$ to user $i$. In making this recommendation CNE did not have access to any meta-data of the users or movies, and neither does ExplaiNE to identify explanations. Yet, we can make use of this meta-data to qualitatively assess whether the explanations make sense.
As an example, we computed the recommendation for the first user (uid=$0$) in the user list (See Table.~\ref{tb:imdb_explanation}). The top recommended movie is `Batman' with genre tags `Action', `Adventure', `Crime', and `Drama'. The genres of the top explanations given by ExplainNE arguably have strongly overlapping genre tags (e.g., all top-$5$ are tagged with `Action'). Moreover, the second-highest ranked explanation is `Batman Forever'.

\begin{table}[t]
  \caption{Recommended movie to user uid=$0$. The top movie recommended by CNE (Batman) is explained through movies already seen by user uid=$0$. The top-ranked explanations have genres that overlap with the recommended movie. \label{tb:imdb_explanation}}
\centering
\setlength{\tabcolsep}{2pt}
  \begin{tabular}{ c  p{34mm}  p{36mm}}
    \multicolumn{1}{c}{$j$} & \multicolumn{1}{c}{Recommendations} & \multicolumn{1}{c}{Genres}\\\hline
    1 & Batman & Action, Adventure, Crime, Drama\\
    2 & E.T. the Extra-Terrestrial & Children's, Drama, Fantasy, Sci-Fi\\
    3 & The Secret of Roan Inish & Adventure \\

    \multicolumn{1}{c}{}&\multicolumn{1}{c}{}&\multicolumn{1}{c}{}\\

    $k$& \multicolumn{1}{c}{Explanations for `Batman'} & \multicolumn{1}{c}{Genres}\\\hline
    1& Supercop & Action, Thriller\\
    2& Batman Forever & Action, Adventure, Comedy, Crime\\
    3& The Crow & Action, Romance, Thriller \\
    4& Full Metal Jacket & Action, Drama, War \\
    5& Young Guns & Action, Comedy, Western \\
  \end{tabular}
	\vskip -0.6cm
\end{table}

More case studies are given in the supplement Sec.~5.
\todo[inline]{Correct, Bo?}
These results suggest that ExplaiNE gives sensible explanations.
The next subsection aims to quantify these findings.

\subsection{Quantitative evaluation\label{sec:exp_quantitative}}

Objectively evaluating the quality of an explanation is conceptually non-trivial,
due to a lack of datasets with ground-truth explanations for LP.
Yet, as we show in this section, it is possible to use metadata to derive reasonable ground truth explanations, and compare with those.

\ptitlenoskip{DBLP network} 
Here, we can construct ground truth explanations for \emph{existing} links (as opposed to \emph{predicted} ones). While this is not the intended use case of ExplaiNE, it is perfectly legitimate and justified here given our intention to objectively validate the quality of the explanations.
Our approach is based on the intuition that a one-time co-author $j$ of a given author $i$ could have been introduced to that author $i$ by another co-author $k$ on the same paper, thus explaining the link $\{i,j\}$. While this will of course not always be true, we postulate that it is sufficiently common for ExplaiNE---providing it works well---to highlight the other co-authors as explanations for the observed link $\{i,j\}$.

\begin{figure*}[t]
\centering
   \includegraphics[trim=0 0 0 0, clip=true, width=\textwidth]{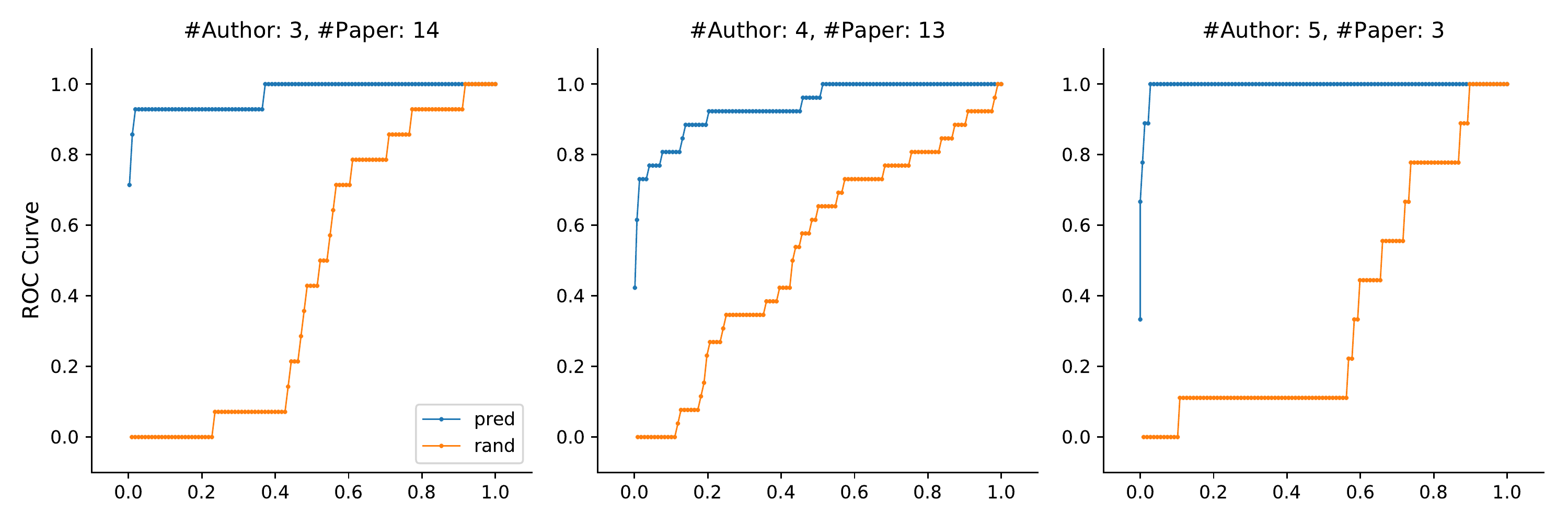}
	\vskip -0.5cm
\caption{ROC curves of co-author predictions for $i=$'Eric P. Xing', with author-list lengths $3$, $4$, and $5$ (orange=rand., blue=ExplaiNE). \label{fig:exp_quant_dblp}}
\end{figure*}

Given an author $i$ and a one-time co-author $j$ of $i$, 
we used ExplaiNE to rank the other co-authors of $i$, from more to less explanatory (according to Eq.~\ref{eq:approx_der}).
We then took the top-$r$ of this ranked list as predicted co-authors on the paper $i$ co-authored with $j$. Based on this, we created a confusion matrix. Clearly, the hardness of this prediction task is different for papers with different numbers of authors. Thus, in order to get a more aggregate assessment, we summed the top-$r$ confusion matrices for all one-time co-authors of node $i$ on papers with a given number of co-authors $L$, and this for different $L$ between 3 and 5. For a given author-list length, the confusion matrices with different $r$ were then used to create precision-recall curves or ROC curves.
Figure~\ref{fig:exp_quant_dblp} shows the ROC curves for Eric P. Xing as node $i$ and three author-list lengths.
For comparison, also ROC curves computed based on a randomly ranked list is shown (as the size of the data is rather small, these are not always close to the diagonal).
ROC curves for other nodes $i$ as well as Precision-Recall curves can be found in the supplement.
\todo[inline]{True, Bo?}
All results indicate that the explanations are remarkably effective at this task, indicating that 
ExplaiNE performs well.

\begin{figure}[t]
  \centering
  \includegraphics[width=0.5\textwidth]{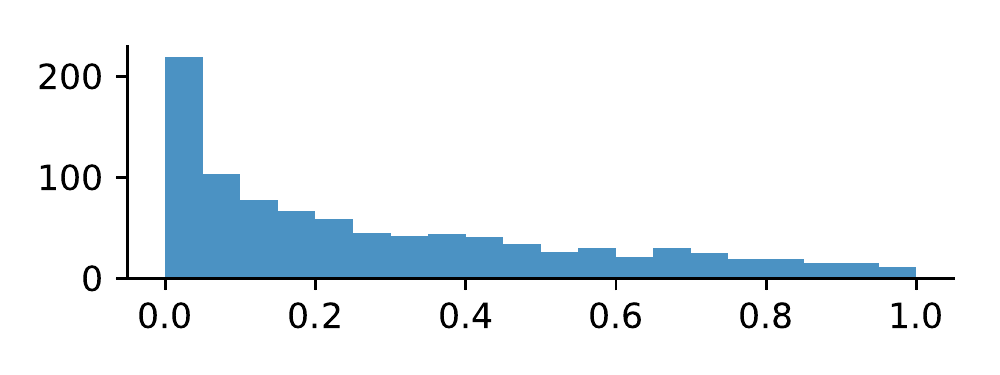}
	\vskip -0.4cm
\caption{$P$-values that indicates the significance of the correlation between the genre recommended and the genres in the explanation. Each $p$-value is computed against $50$ random explanations. Those explanations are drawn from user's watched movies. The empirical distribution has Kolmogorov-Smirnov test statistic $0.32$ and a $p$-value that is numerically $0.0$ against uniform distribution. This shows the significance of positive correlation between the recommended movies and the explanations made by ExplaiNE. \label{fig:exp_quant_imdb}}
\end{figure}

\ptitlenoskip{MovieLens network} A good explanation $k$ of a predicted link between a movie-user pair $\{i,j\}$ should arguably have a similar list of genres as $j$. To test this, we computed the top-$5$ explanations for user $i$ and her top recommended movie $j$. Then we averaged the Jaccard similarity between the set of genres for movie $j$ and the set of genres of each of the $5$ explanations. To assess the significance of this average, we computed an empirical $p$-value for it by randomly sampling $50$ sets of $5$ `explanations' drawn from the watched movies of $i$, resulting in $50$ random average Jaccard similarities to compare with the one obtained by ExplaiNE. Thus we obtained an empirical $p$-value for each user $i$, indicating the significance of the overlap between the set of genres of the recommended movie $j$ and the top-$5$ explanations.
A histogram of these $p$-values is shown in Fig.~\ref{fig:exp_quant_imdb}. While $p$-values are uniformly distributed under the null hypothesis that the explanations have genres unrelated to those of $j$, here this is not the case---indicating the null hypothesis is false. A Kolmogorov-Smirnoff test indeed shows an extremely high significance ($p$-value numerically $0$).

\subsection{Scalability and runtime\label{sec:exp_runtime}}

To address \textbf{Q3}, we measured the runtime of exact and approximated ExplaiNE when computing $\frac{\partial g^*_{ij}}{\partial a_{ik}}(\hmA)$ for all $k\not\in\{i,j\}$, as per Eqs.~(\ref{eq:exact_der}) and (\ref{eq:approx_der}), on average over random pairs of nodes $\{i,j\}$. The runtime was measured on a PC with quad-core 2.7GHz Intel Core i5 and 16GB 1600MHz DDR3 RAM. Table~\ref{tb:exp_runtime}, shows that approximated ExplaiNE is efficient and applicable to large networks with higher dimensionality, while exact ExplaiNE is not.

\begin{table} \centering
  \caption{Average runtime (in sec., $10$ trials) of exact and approximated ExplaiNE in computing the explanations for a random pair of nodes $\{i,j\}$. 
Note that the exact method also has substantial memory cost: $13.1$ Gb for MovieLens and on DBLP we went out of memory. On MovieLens, the time was computed only for one $k$, and multiplied by $n-2$ to get an estimated total time for all $k$.
\label{tb:exp_runtime}}
  \begin{tabular}{c c c c c}
    Network & $\#$nodes & dim & time exact & time approx \\\hline
    Karate & $34$ & $2$ & $0.03$ & $1.8\mathrm{e}{-4}$ \\
    GoT & $796$ & $2$ & $64.1$ & $4.1\mathrm{e}{-4}$ \\
    GoT & $796$ & $8$ & $1490.2$ & $9.8\mathrm{e}{-4}$ \\
    MovieLens & $2625$ & $16$ & $\sim1.63\mathrm{e}{6}$ & $6.8\mathrm{e}{-3}$\\
    DBLP & $23359$ & $32$ & --- & $0.02$
  \end{tabular}
	\vskip -0.8cm
\end{table}

\section{Related Work\label{sec:relatedwork}}

LP, as an important network analysis task, has recently been extensively studied in the NE literature \citep{hamilton2017representation, cui2018survey}. By embedding the nodes in a vector space, the link prediction task can be addressed using traditional machine learning (ML) methods. This has led to new and accurate approaches for LP \citep{grover2016node2vec, kang2018conditional}, but at the expense of explainability. 

In parallel, the importance of accountability of AI has sparked growing research interest in interpretable ML. Approaches to interpretable machine learning research can be categorized into model-based and post-hoc approaches \citep{du2018techniques, murdoch2019interpretable}. The first category focuses on incorporating interpretability (e.g., sparsity) while constructing the ML model. ExplaiNE belongs to the second category of interpretable ML methods: it is a post-hoc method that focuses on interpreting the local structure of ML models (here, NE models).
The most strongly related work (although not for LP) are \citet{ribeiro2016should} and \citet{lundberg2017unified}, who provide a model-agnostic explanation via local approximation of the target model. More closely related, \citet{simonyan2013deep} and \citet{koh2017understanding} propose to compute the gradient of the loss function of a (black-box) model with respect to the input to gauge the relevance of the input features. The first of these computes the gradient using back-propagation, while the second approximates the gradient using a Taylor series expansion.

ExplaiNE is the first generic approach (and, as far as we know, the first approach at all) for explaining link predictions based on a NE.
Moreover, to the best of our knowledge, ExplaiNE is the first method that uses the implicit function theorem for explainability. This proved to be a crucial element for computing the gradient of the link probability w.r.t. the network structure, as it allowed us to rigorously track the optimal embedding given an infinitesimal change in the input network. We believe this theorem can prove valuable also for other tasks, particularly those where an intermediate representation is obtained by optimizing an unsupervised objective function (e.g., an autoencoder), to be fed into a subsequent model that is trained in a supervised manner.

\section{Conclusions\label{sec:conclusions}}

Link Prediction (LP) in networks is an important task, with applications to social networks, recommenders, and knowledge graphs. State-of-the-art approaches are based on first embedding the nodes in a vector space, followed by a LP step.
Unfortunately, while accurate, these approaches offer no insight in their predictions.
To remedy this, we introduced ExplaiNE, a generic approach to explain LPs based on Network Embeddings (NEs) in terms of existing links in the network.
ExplaiNE is applicable for a wide range of NE methods.
We applied it to CNE, a state-of-the-art NE method, and outlined how it can be applied for a wide range of other NE methods. Extensive qualitative and quantitative evaluations show the usefulness of ExplaiNE, and its ability to scale to large networks.

In the future we aim to develop ExplaiNE for other NE methods, apply it to recommender systems, and extend it to offer explanations in terms of the presence of dense communities or other larger substructures in the network.

\subsubsection*{Acknowledgments}
The research leading to these results has received funding from the European Research Council under the European Union's Seventh Framework Programme (FP7/2007-2013) / ERC Grant Agreement no. 615517, from the FWO (project no. G091017N, G0F9816N), and from the European Union's Horizon 2020 research and innovation programme and the FWO under the Marie Sklodowska-Curie Grant Agreement no. 665501.

\bibliography{paper}

\begin{thebibliography}{24}
\providecommand{\natexlab}[1]{#1}
\providecommand{\url}[1]{\texttt{#1}}
\expandafter\ifx\csname urlstyle\endcsname\relax
  \providecommand{\doi}[1]{doi: #1}\else
  \providecommand{\doi}{doi: \begingroup \urlstyle{rm}\Url}\fi

\bibitem[Adriaens et~al.(2017)Adriaens, Lijffijt, and
  De~Bie]{adriaens2017subjectively}
Adriaens, F., Lijffijt, J., and De~Bie, T.
\newblock Subjectively interesting connecting trees.
\newblock In \emph{Joint European Conference on Machine Learning and Knowledge
  Discovery in Databases}, pp.\  53--69. Springer, 2017.

\bibitem[Chiang(1984)]{chiang1984fundamental}
Chiang, A.~C.
\newblock \emph{Fundamental methods of mathematical economics}.
\newblock Aukland (New Zealand) McGraw-Hill, 1984.

\bibitem[Cui et~al.(2018)Cui, Wang, Pei, and Zhu]{cui2018survey}
Cui, P., Wang, X., Pei, J., and Zhu, W.
\newblock A survey on network embedding.
\newblock \emph{IEEE Transactions on Knowledge and Data Engineering}, 2018.

\bibitem[Du et~al.(2018)Du, Liu, and Hu]{du2018techniques}
Du, M., Liu, N., and Hu, X.
\newblock Techniques for interpretable machine learning.
\newblock \emph{arXiv preprint arXiv:1808.00033}, 2018.

\bibitem[Grover \& Leskovec(2016)Grover and Leskovec]{grover2016node2vec}
Grover, A. and Leskovec, J.
\newblock node2vec: Scalable feature learning for networks.
\newblock In \emph{Proceedings of the 22nd ACM SIGKDD international conference
  on Knowledge discovery and data mining}, pp.\  855--864. ACM, 2016.

\bibitem[Hamilton et~al.(2017)Hamilton, Ying, and
  Leskovec]{hamilton2017representation}
Hamilton, W.~L., Ying, R., and Leskovec, J.
\newblock Representation learning on graphs: Methods and applications.
\newblock \emph{arXiv preprint arXiv:1709.05584}, 2017.

\bibitem[Harper \& Konstan(2016)Harper and Konstan]{harper2016movielens}
Harper, F.~M. and Konstan, J.~A.
\newblock The movielens datasets: History and context.
\newblock \emph{Acm transactions on interactive intelligent systems (tiis)},
  5\penalty0 (4):\penalty0 19, 2016.

\bibitem[Kang et~al.(2019)Kang, Lijffijt, and {De Bie}]{kang2018conditional}
Kang, B., Lijffijt, J., and {De Bie}, T.
\newblock Conditional network embeddings.
\newblock In \emph{International Conference on Learning Representations}, 2019.
\newblock URL \url{https://openreview.net/forum?id=ryepUj0qtX}.

\bibitem[Koh \& Liang(2017)Koh and Liang]{koh2017understanding}
Koh, P.~W. and Liang, P.
\newblock Understanding black-box predictions via influence functions.
\newblock In \emph{International Conference on Machine Learning}, pp.\
  1885--1894, 2017.

\bibitem[Leone et~al.(1961)Leone, Nelson, and Nottingham]{leone1961folded}
Leone, F., Nelson, L., and Nottingham, R.
\newblock The folded normal distribution.
\newblock \emph{Technometrics}, 3\penalty0 (4):\penalty0 543--550, 1961.

\bibitem[Levy \& Goldberg(2014)Levy and Goldberg]{levy2014neural}
Levy, O. and Goldberg, Y.
\newblock Neural word embedding as implicit matrix factorization.
\newblock In \emph{Advances in neural information processing systems}, pp.\
  2177--2185, 2014.

\bibitem[Lundberg \& Lee(2017)Lundberg and Lee]{lundberg2017unified}
Lundberg, S.~M. and Lee, S.-I.
\newblock A unified approach to interpreting model predictions.
\newblock In \emph{Advances in Neural Information Processing Systems}, pp.\
  4765--4774, 2017.

\bibitem[Mart{\'\i}nez et~al.(2017)Mart{\'\i}nez, Berzal, and
  Cubero]{martinez2017survey}
Mart{\'\i}nez, V., Berzal, F., and Cubero, J.-C.
\newblock A survey of link prediction in complex networks.
\newblock \emph{ACM Computing Surveys (CSUR)}, 49\penalty0 (4):\penalty0 69,
  2017.

\bibitem[Mikolov et~al.(2013)Mikolov, Sutskever, Chen, Corrado, and
  Dean]{mikolov2013distributed}
Mikolov, T., Sutskever, I., Chen, K., Corrado, G.~S., and Dean, J.
\newblock Distributed representations of words and phrases and their
  compositionality.
\newblock In \emph{Advances in neural information processing systems}, pp.\
  3111--3119, 2013.

\bibitem[Murdoch et~al.(2019)Murdoch, Singh, Kumbier, Abbasi-Asl, and
  Yu]{murdoch2019interpretable}
Murdoch, W.~J., Singh, C., Kumbier, K., Abbasi-Asl, R., and Yu, B.
\newblock Interpretable machine learning: definitions, methods, and
  applications.
\newblock \emph{arXiv preprint arXiv:1901.04592}, 2019.

\bibitem[Perozzi et~al.(2014)Perozzi, Al-Rfou, and Skiena]{perozzi2014deepwalk}
Perozzi, B., Al-Rfou, R., and Skiena, S.
\newblock Deepwalk: Online learning of social representations.
\newblock In \emph{Proceedings of the 20th ACM SIGKDD international conference
  on Knowledge discovery and data mining}, pp.\  701--710. ACM, 2014.

\bibitem[Qiu et~al.(2018)Qiu, Dong, Ma, Li, Wang, and Tang]{qiu2018network}
Qiu, J., Dong, Y., Ma, H., Li, J., Wang, K., and Tang, J.
\newblock Network embedding as matrix factorization: Unifying deepwalk, line,
  pte, and node2vec.
\newblock In \emph{Proceedings of the Eleventh ACM International Conference on
  Web Search and Data Mining}, pp.\  459--467. ACM, 2018.

\bibitem[Ribeiro et~al.(2016)Ribeiro, Singh, and Guestrin]{ribeiro2016should}
Ribeiro, M.~T., Singh, S., and Guestrin, C.
\newblock Why should i trust you?: Explaining the predictions of any
  classifier.
\newblock In \emph{Proceedings of the 22nd ACM SIGKDD international conference
  on knowledge discovery and data mining}, pp.\  1135--1144. ACM, 2016.

\bibitem[Simonyan et~al.(2013)Simonyan, Vedaldi, and
  Zisserman]{simonyan2013deep}
Simonyan, K., Vedaldi, A., and Zisserman, A.
\newblock Deep inside convolutional networks: Visualising image classification
  models and saliency maps.
\newblock \emph{arXiv preprint arXiv:1312.6034}, 2013.

\bibitem[Tang et~al.(2008)Tang, Zhang, Yao, Li, Zhang, and
  Su]{tang2008arnetminer}
Tang, J., Zhang, J., Yao, L., Li, J., Zhang, L., and Su, Z.
\newblock Arnetminer: extraction and mining of academic social networks.
\newblock In \emph{Proceedings of the 14th ACM SIGKDD international conference
  on Knowledge discovery and data mining}, pp.\  990--998. ACM, 2008.

\bibitem[Tang et~al.(2015{\natexlab{a}})Tang, Qu, and Mei]{tang2015pte}
Tang, J., Qu, M., and Mei, Q.
\newblock Pte: Predictive text embedding through large-scale heterogeneous text
  networks.
\newblock In \emph{Proceedings of the 21th ACM SIGKDD International Conference
  on Knowledge Discovery and Data Mining}, pp.\  1165--1174. ACM,
  2015{\natexlab{a}}.

\bibitem[Tang et~al.(2015{\natexlab{b}})Tang, Qu, Wang, Zhang, Yan, and
  Mei]{tang2015line}
Tang, J., Qu, M., Wang, M., Zhang, M., Yan, J., and Mei, Q.
\newblock Line: Large-scale information network embedding.
\newblock In \emph{Proceedings of the 24th International Conference on World
  Wide Web}, pp.\  1067--1077. International World Wide Web Conferences
  Steering Committee, 2015{\natexlab{b}}.

\bibitem[van Leeuwen et~al.(2016)van Leeuwen, De~Bie, Spyropoulou, and
  Mesnage]{van2016subjective}
van Leeuwen, M., De~Bie, T., Spyropoulou, E., and Mesnage, C.
\newblock Subjective interestingness of subgraph patterns.
\newblock \emph{Machine Learning}, 105\penalty0 (1):\penalty0 41--75, 2016.

\bibitem[Zachary(1977)]{zachary1977information}
Zachary, W.~W.
\newblock An information flow model for conflict and fission in small groups.
\newblock \emph{Journal of anthropological research}, 33\penalty0 (4):\penalty0
  452--473, 1977.

\end{thebibliography}
\bibliographystyle{icml2019}

\onecolumn
\icmltitle{Supplementary materials for ExplaiNE}







\icmlkeywords{Explanability, Network Embedding}

\vskip 0.3in



\setcounter{section}{0}
\setcounter{equation}{0}
\section{Approximated ExplaiNE for Conditional Network Embedding}
We now apply approximated ExplaiNE approach to Conditional Network Embedding. Our goal is to compute gradient of predicted link probability $g_{ij}^*$ with respect to the change of link $a_{ik}$ (main paper Eq.7), namely:

\begin{align}\label{eq:supp_approx_der}
\frac{\partial g_{ij}^*}{\partial a_{ik}}(\hmA) &= -\nabla_{{\vx_i}}g_{ij}\left(\Xopt\right)^T\cdot
\mH_i^{-1} \cdot\frac{\partial \hat\mF_i}{\partial a_{ik}}(\hmA,\hat\vx_i^*).
\end{align}

In order to compute the derivative, we first drive the optimality conditions for CNE. Recall the objective function $\cL$ in CNE is log-probability:
\begin{align*}
\cL(\hmA,\mX)=\log(P(\hmA|\mX)) &= \sum_{\{i,j\}:\ha_{ij}=1} \log{P_{ij}(a_{ij}=1|\mX)} \\
&+ \sum_{\{i,j\}:\ha_{ij}=0} \log{P_{ij}(a_{ij}=0|\mX)}.
\end{align*}
where the link probabilities conditioned on the embedding are defined as:
\begin{align*}
P(a_{ij}=1|\mX) &=
\frac{P_{\hmA,ij}\cN_{+,\sigma_1}(\|\vx_i-\vx_j\|)}{P_{\hmA,ij}\cN_{+,\sigma_1}(\|\vx_i-\vx_j\|) + (1-P_{\hmA,ij})\cN_{+,\sigma_2}(\|\vx_i-\vx_j\|)},\\
P(a_{ij}=0|\mX) &= 1-P(a_{ij}=1|\mX)
\end{align*}
where $\cN_{+,\sigma}$ denotes a half-Normal distribution \citep{leone1961folded} with spread parameter $\sigma$, where $\sigma_2>\sigma_1=1$, and where $P_{\hmA,ij}$ is a prior probability for a link to exist between nodes $i$ and $j$ as inferred from the degrees of the nodes (or based on other types of information about the structure of the network), derived as explained in \citet{adriaens2017subjectively, van2016subjective}.

Denoting $\gamma = \frac{1}{\sigma_1^2} - \frac{1}{\sigma_2^2}$, and $\hat{P}^*_{ij}\triangleq g_{ij}^*(\hmA)=P(a_{ij} = 1 | \Xopt)$ (the probability of a link between $i$ and $j$ given the optimal embedding $\Xopt$ for $\hmA$), we can now derive the three terms in Eq.~\ref{eq:supp_approx_der}.

\ptitlenoskip{First term.}
First term is derived by taking the gradient of predicted link probability $g_{ij}\left(\Xopt\right)$ with respect to $\vx_i$:
\begin{align}\label{eq:supp_approx_der_first}
\nabla_{\vx_i}g_{ij}\left(\Xopt\right)&=\nabla_{\vx_i} P(a_{ij}=1|\mX) \nonumber\\
&=\nabla_{\vx_i} \left(\frac{P_{\hmA, ij}\cN_{+,\sigma_1}(\|\vx_i^*-\vx_j^*\|)}{P_{\hmA, ij}\cN_{+,\sigma_1}(\|\vx_i^*-\vx_j^*\|)+(1-P_{\hmA, ij})\cN_{+,\sigma_2}(\|\vx_i^*-\vx_j^*\|)}\right) \nonumber\\
&= \frac{\nabla_{\vx_i} \left(P_{\hmA, ij}\cN_{+,\sigma_1}(\|\vx_i^*-\vx_j^*\|)\right)
\left(P_{\hmA, ij}\cN_{+,\sigma_1}(\|\vx_i^*-\vx_j^*\|)+(1-P_{\hmA, ij})\cN_{+,\sigma_2}(\|\vx_i^*-\vx_j^*\|)\right)}%
{\left(P_{\hmA, ij}\cN_{+,\sigma_1}(\|\vx_i^*-\vx_j^*\|)+(1-P_{\hmA, ij})\cN_{+,\sigma_2}(\|\vx_i^*-\vx_j^*\|)\right)^2}\nonumber\\
&- \frac{\left(P_{\hmA, ij}\cN_{+,\sigma_1}(\|\vx_i^*-\vx_j^*\|)\right)
\nabla_{\vx_i} \left(P_{\hmA, ij}\cN_{+,\sigma_1}(\|\vx_i^*-\vx_j^*\|)+(1-P_{\hmA, ij})\cN_{+,\sigma_2}(\|\vx_i^*-\vx_j^*\|)\right)}%
{\left(P_{\hmA, ij}\cN_{+,\sigma_1}(\|\vx_i^*-\vx_j^*\|)+(1-P_{\hmA, ij})\cN_{+,\sigma_2}(\|\vx_i^*-\vx_j^*\|)\right)^2}\nonumber\\
&= \frac{\frac{-1}{\sigma_1^2}(\vx_i^*-\vx_j^*)\left(P_{\hmA, ij}\cN_{+,\sigma_1}(\|\vx_i^*-\vx_j^*\|)\right)
\left(P_{\hmA, ij}\cN_{+,\sigma_1}(\|\vx_i^*-\vx_j^*\|)+(1-P_{\hmA, ij})\cN_{+,\sigma_2}(\|\vx_i^*-\vx_j^*\|)\right)}%
{\left(P_{\hmA, ij}\cN_{+,\sigma_1}(\|\vx_i^*-\vx_j^*\|)+(1-P_{\hmA, ij})\cN_{+,\sigma_2}(\|\vx_i^*-\vx_j^*\|)\right)^2}\nonumber\\
&- \frac{\left(P_{\hmA, ij}\cN_{+,\sigma_1}(\|\vx_i^*-\vx_j^*\|)\right)
\left(\frac{-1}{\sigma_1^2}(\vx_i^*-\vx_j^*)\left(P_{\hmA, ij}\cN_{+,\sigma_1}(\|\vx_i^*-\vx_j^*\|)\right)
+\frac{-1}{\sigma_2^2}(1-P_{\hmA, ij})\cN_{+,\sigma_2}(\|\vx_i^*-\vx_j^*\|)\right)}%
{\left(P_{\hmA, ij}\cN_{+,\sigma_1}(\|\vx_i^*-\vx_j^*\|)+(1-P_{\hmA, ij})\cN_{+,\sigma_2}(\|\vx_i^*-\vx_j^*\|)\right)^2}\nonumber\\
&= -\gamma\frac{P_{\hmA, ij}(1-P_{\hmA, ij})\cN_{+,\sigma_1}(\|\vx_i^*-\vx_j^*\|)\cN_{+,\sigma_2}(\|\vx_i^*-\vx_j^*\|)}%
{\left(P_{\hmA, ij}\cN_{+,\sigma_1}(\|\vx_i^*-\vx_j^*\|)+(1-P_{\hmA, ij})\cN_{+,\sigma_2}(\|\vx_i^*-\vx_j^*\|)\right)^2}\nonumber\\
&= -\gamma(\vx_i^*-\vx_j^*)P(a_{ij}=1|\Xopt)P(a_{ij}=0|\Xopt)\nonumber\\
&= -\gamma(\vx_i^*-\vx_j^*)\Popt_{ij}(1-\Popt_{ij})
\end{align}

\ptitlenoskip{Second term}
The Hessian $\mH_i$ the gradient of $\hat\mF_i(\hmA, \hat\vx_i^*)$ w.r.t. $\vx_i^*$. Recall the optimality condition:
The optimality condition in approximated ExplaiNE can be derived as:
\begin{align}\label{eq:supp_approx_optimality}
\hat\mF_i(\hmA, \hat\vx_i^*) &= \nabla_{\vx_i^*}\log(P(\hmA|\Xopt)) \nonumber\\
&= \gamma\sum_{j\neq i} (\hat\vx_i^*-\hat\vx_j^*)\left(P\left(a_{ij}=1|\Xopt\right)-\ha_{ij}\right) \nonumber\\
&= \mathbf{0}.
\end{align}
Based on this condition we derive the Hessian:
\begin{align}
\mH_i &=\nabla_{\vx_i}\hat\mF_i(\hmA, \hat\vx_i^*) \nonumber\\
&=\nabla_{\vx_i}\left(\gamma\sum_{j\neq i} (\hat\vx_i^*-\hat\vx_j^*)\left(P\left(a_{ij}=1|\Xopt\right)-\ha_{ij}\right)\right)\nonumber\\
&=\gamma\left(
\mI\sum_{j\neq i} \left(P_{ij}^*-\ha_{ij}\right)
-\gamma\sum_{j\neq i} (\vx_i^*-\vx_j^*)(\vx_i^*-\vx_j^*)'P_{ij}^*(1-P_{ij}^*)\right).
\end{align}
The last line follows similar derivation of Eq.~\ref{eq:supp_approx_der_first}.

\ptitlenoskip{Third term}
Now compute the gradient with respect to $a_{ik}$ from optimality condition (Eq.~\ref{eq:supp_approx_optimality}):
\begin{align}
\frac{\partial\hat\mF_i}{\partial \ha_{ik}}(\hmA, \hat\vx_i^*) &= \frac{\partial}{\partial \ha_{ik}} \left(\gamma\sum_{j\neq i} (\hat\vx_i^*-\hat\vx_j^*)\left(P\left(a_{ij}=1|\Xopt\right)-\ha_{ij}\right)\right) \nonumber\\
&=\gamma(\vx_k^* - \vx_i^*).
\end{align}
Putting things together, we have
\begin{align}
  &\frac{\partial g_{ij}^*}{\partial a_{ik}}(\hmA) \nonumber\\
	&= (\vx_i^*-\vx_j^*)'\cdot\left(\frac{-\mH_i}{\gamma^2\hat{P}^*_{ij}(1-\hat{P}^*_{ij})}\right)^{-1}\cdot(\vx_i^* - \vx_k^*)
\end{align}

\section{Exact ExplaiNE for Conditional Network Embedding}
We now apply exact ExplaiNE approach to Conditional Network Embedding (CNE). Our goal is to compute the change (i.e., gradient) of link probability of a node paire $\{i,j\}$ with respect to node-pairs $\{k,l\}$ ($k\neq l$ and $\{k,l\}\neq \{i,j\}$), i.e., the explanations. Applying the chain rule and implicit function theorem, the exact ExplaiNE computes this gradient as (main paper Eq.4):
\begin{align}\label{eq:supp_eaxct_gradient}
\frac{\partial g_{ij}^*}{\partial a_{kl}}(\hmA) &=-\nabla_{{\mX}}g_{ij}\left(\Xopt\right)^T\cdot\mH^{-1}\cdot\frac{\partial\mF}{\partial a_{kl}}(\hmA, \Xopt).
\end{align}
Now let's derive each term in Eq.~\ref{eq:supp_eaxct_gradient} for CNE.

\ptitlenoskip{First term} Because in CNE the link function $g_{ij}^*(\hmA)$ only depends on $\vx_i$ and $\vx_j$, the gradient $\nabla_{{\mX}}g_{ij}\left(\Xopt\right)$ is a $nd\times 1$ vector with $2d$ non-zero entries. Formally:
\begin{equation}
\nabla_{{\mX}}g_{ij}\left(\Xopt\right) =
  \begin{bmatrix}
    \mathbf{0} \\ \vdots \\ \nabla_{{\vx_i}}g_{ij}\left(\Xopt\right) \\ \vdots \\ \nabla_{{\vx_j}}g_{ij}\left(\Xopt\right)  \\ \vdots \\ \mathbf{0}
  \end{bmatrix}
\end{equation}
where
\begin{equation*}
  \nabla_{{\vx_i}}g_{ij}\left(\Xopt\right) = -\gamma(\vx_i^* - \vx_j^*)\Popt_{ij}(1-\Popt_{ij}),
\end{equation*}
and
\begin{equation*}
  \nabla_{{\vx_j}}g_{ij}\left(\Xopt\right) = -\gamma(\vx_j^* - \vx_i^*)\Popt_{ij}(1-\Popt_{ij}).
\end{equation*}

\ptitlenoskip{Second term}
We drive the Hessian $\mH$ ($nd\times nd$) as:
\begin{align}
\mH &= \nabla_{\mX} \mF(\hmA, \mX^*) \nonumber\\
&=\begin{bmatrix}
    & \ddots & \\
    & & \nabla_{\vx_i^*} \mF_i(\hmA, \vx_i^*) & \dots & \nabla_{\vx_i^*}\mF_j(\hmA, \vx_j^*) &  \\
    & & \vdots & \ddots & \vdots & \\
    & & \nabla_{\vx_j^*} \mF_i(\hmA, \vx_i^*) & \dots & \nabla_{\vx_j^*}\mF_j(\hmA, \vx_j^*) & \\
    & & & & & \ddots &
  \end{bmatrix},
\end{align}
where
\begin{equation*}
\nabla_{\vx_i^*}\mF_i(\hmA, \vx_i^*)=\gamma\mI\sum_{j\neq i}(\Popt_{ij} - \ha_{ij}) -\gamma^2\sum_{j\neq i}(\vx_i^* - \vx_j^*)(\vx_i^* - \vx_j^*)' \Popt_{ij}(1-\Popt_{ij})
\end{equation*}
and
\begin{equation*}
\nabla_{\vx_j^*}\mF_j(\hmA, \vx_i^*)=-\gamma\mI(\Popt_{ij} - \ha_{ij}) + \gamma^2(\vx_i^* - \vx_j^*)(\vx_i^* - \vx_j^*)' \Popt_{ij}(1-\Popt_{ij})
\end{equation*}

\ptitlenoskip{Third term} Finally, we can compute derivative $\frac{\partial\mF}{\partial a_{kl}}(\hmA, \Xopt)$ ($nd\times 1$) as:
\begin{equation}
\frac{\partial\mF}{\partial a_{kl}}(\hmA, \Xopt) =
\begin{bmatrix}
  \mathbf{0} \\ \vdots \\ \gamma(\vx_k^* - \vx_l^*) \\ \vdots \\ \gamma(\vx_l^* - \vx_k^*) \\ \vdots \\ \mathbf{0}
\end{bmatrix}
\end{equation}
Putting all together, we have gradient of link probability $g_{ij}^*(\hmA)$ with respect to a link $a_{kl}$:

\begin{align*}\label{eq:supp_eaxct_gradient}
\frac{\partial g_{ij}^*}{\partial a_{kl}}(\hmA) &=-\nabla_{{\mX}}g_{ij}\left(\Xopt\right)^T\cdot\mH^{-1}\cdot\frac{\partial\mF}{\partial a_{kl}}(\hmA, \Xopt) \\
  &=\begin{bmatrix}
    \mathbf{0} \\ \vdots \\ \nabla_{{\vx_i}}g_{ij}\left(\Xopt\right) \\ \vdots \\ \nabla_{{\vx_j}}g_{ij}\left(\Xopt\right)  \\ \vdots \\ \mathbf{0}
  \end{bmatrix}'\cdot \begin{bmatrix}
      & \ddots & \\
      & & -\nabla_{\vx_i^*} \mF_i(\hmA, \vx_i^*) & \dots & -\nabla_{\vx_i^*}\mF_j(\hmA, \vx_j^*) &  \\
      & & \vdots & \ddots & \vdots & \\
      & & -\nabla_{\vx_j^*} \mF_i(\hmA, \vx_i^*) & \dots & -\nabla_{\vx_j^*}\mF_j(\hmA, \vx_j^*) & \\
      & & & & & \ddots &
    \end{bmatrix}^{-1} \begin{bmatrix}
      \mathbf{0} \\ \vdots \\ \gamma(\vx_k^* - \vx_l^*) \\ \vdots \\ \gamma(\vx_l^* - \vx_k^*) \\ \vdots \\ \mathbf{0}
    \end{bmatrix}
\end{align*}

\section{Approximated ExplaiNE for LINE}
LINE \citep{tang2015line} computes embeddings that approximate both 1st and 2nd order proximity. The model consists of two objective functions. The first one measures the distance (KL) between the model and empirical distribution of the 1st proximity measure. Similarly the second objective function measures the distance (KL) between the model and empirical distribution of the 2nd proximity measure. Optimizing over the objective functions thus gives two embeddings for an input word, one for each type of proximities. Then the embeddings are simply concatenated and used as the final node embedding. Usually the context embedding in the 2nd proximity approximation are not used in the practice, similar to the situation in the language modeling (e.g., skip-gram with negative sampling a.k.a SGNS).

Similar to CNE, we assume the given network is unweighted, undirected. Explaining the first order proximity embedding (LINE 1st) is simple, thus our derivation assumes the link prediction is based solely on the LINE 2nd embedding, adopt the reasoning in the works \citep{levy2014neural, qiu2018network}. For LINE 2nd, the probability of encountering ``context'' node $v_j$ given node $v_i$ is defined as,
\begin{equation*}
  p(v_j|v_i) = \frac{\exp(\vx_i'\vy_j)}{\sum_{k}\exp(\vx_i' \vy_k)}
\end{equation*}
where $\vx_i$ denotes the embedding of a target node $i$ and $\vy_j$ denotes the embedding of context node $j$.

Then the 2nd proximity objective reads:
\begin{equation*}
O = \sum_{i\in V} d_i \text{KL}\left(\hat{p}(\cdot|v_i) \| p(\cdot|v_i)\right)
\end{equation*}
where $d_i = \sum_j a_{ij}$ weights the importance of $i$-th node, and empirical distribution $\hat{p}(\cdot | v_i)$ is defined as $\hat{p}(v_j | v_i) = \frac{a_{ij}}{d_i}$. Carrying out the KL-divergence, the objective function can be further expressed as:
\begin{equation*}
O = \sum_{(i,j) \in E}a_{ij} \log p(v_j|v_i)
\end{equation*}

Because the normalization factor of the distribution is computationally expensive, the objective function is approximated using negative sampling \citep{mikolov2013distributed}:
\begin{equation*}
\cL = \sum_{(i,j) \in E} a_{ij}\log g(\vx_i'\vy_j) + b\mathbb{E}_{j'\sim P_N}\left[\log g\left(-\vx_i'\vy_{j'}\right)\right]
\end{equation*}
where $b$ is the number of negative samples, $g(z)$ is the sigmoid function. The expectation term can be expressed as
\begin{equation*}
  \mathbb{E}_{j'\sim P_N}\left[\log g\left(-\vx_i'\vy_{j'}\right)\right] = \sum_{j'} \frac{d_{j'}}{\text{vol}(G)}\log g\left(-\vx_i'\vy_{j'}\right)
\end{equation*}
Then the objective function of LINE (2nd) can be expressed as:
\begin{equation*}
\cL = \sum_{(i,j) \in E} a_{ij} \log g(\vx_i'\vy_j) + b\frac{d_id_j}{\text{vol}(G)}\log g(-\vx_i'\vy_j)
\end{equation*}
Denote $g^+_{ij} = g(\vx_i'\vy_j)$ and $g^-_{ij}=g(-\vx_i'\vy_j)$ we can compute the gradient of $\cL$ with respect to $\vx_i$ as:
\begin{equation*}
  \nabla_{\vx_i} \cL = (\sum_j a_{ij}g^-_{ij} - b\frac{d_id_j}{\text{vol}(G)} g^+_{ij})\cdot \vy_j
\end{equation*}
Equating the gradient to zero we have implicit function given by optimal embedding $\vx_i^*$ and $\vy_j^*$:
\begin{align*}
  \mF_i(\hmA, \hat\vx_i) = \nabla_{\vx_i} L_{ij} &= (\sum_j \ha_{ij} g^-_{ij} - b\frac{d_id_j}{\text{vol}(G)} g^+_{ij})\cdot \vy_j^*\\
  &=\mathbf{0}
\end{align*}

Our goal is to explain the predicted link probability $g_{ij}^*$ with respect the change of an edge $\ha_{kl}$. Assume the function function $g$ to be the logistic regression classifier (with parameter $\vw$ and $b$) trained on edge embedding that combines two node embedding using Hadamard operator (i.e., element wise product denoted as $\vx_i^*\circ\vx_j^*$) \citep{grover2016node2vec}:
\begin{equation}
  g_{ij}^*(\Xopt) = \frac{1}{1+e^{-(\vw'(\vx_i^*\circ\vx_j^*)+b)}}
\end{equation}

Our goal is compute gradient:
\begin{align}\label{eq:supp_approx_line_der}
\frac{\partial g_{ij}^*}{\partial a_{ik}}(\Xopt) &= -\nabla_{{\vx_i}}g_{ij}\left(\Xopt\right)^T\cdot
\mH_i^{-1} \cdot\frac{\partial \hat\mF_i}{\partial a_{ik}}(\hmA,\hat\vx_i^*).
\end{align}

First term in the derivative:
\begin{equation*}
\nabla_{\vx_i} g_{ij}^*(\Xopt) = g_{ij}^*(1-g_{ij}^*)\cdot(\vw\circ\vx_j^*)
\end{equation*}

Second term in the derivative:
\begin{align*}
\mH_i &= \nabla_{\vx_i}\mF_i(\hmA,\hat\vx_i^*) \\
&= \nabla_{\vx_i}\left((\sum_j \ha_{ij}g^-_{ij} - b\frac{d_id_j}{\text{vol}(G)} g^+_{ij})\cdot \vy_j^*\right) \nonumber \\
&= \sum_j -\ha_{ij}g^-_{ij}(1-g^-_{ij})\vy_j^*(\vy_j^*)' - b\frac{d_id_j}{\text{vol}(G)}g^+_{ij}(1-g^+_{ij})\vy_j^*(\vy_j^*)' \nonumber \\
&=-\sum_j\left(\ha_{ij} + \frac{d_id_j}{\text{vol}(G)}\right)g^+_{ij}g^-_{ij}\vy_j^*(\vy_j^*)'
\end{align*}

The third term reads:
\begin{align*}
\frac{\partial\mF_i}{\partial\ha_{ik}}(\hmA,\hat\vx_i^*) &= \frac{\partial}{\partial\ha_{ik}} \left(\sum_j \ha_{ij} g^-_{ij}\vy_j^* - b\frac{d_id_j}{\text{vol}(G)} g^+_{ij}\cdot \vy_j^*\right) \nonumber \\
&=g^-_{ik}\vy_k^* - b\left(\frac{\partial}{\partial\ha_{ij}}\frac{d_i}{\text{vol}(G)}\cdot\sum_jd_jg^+_{ij}\vy_j^*\right) \nonumber\\
&=g^-_{ik}\vy_k^* - b\left(\frac{\text{vol}(G) - d_i}{\text{vol}(G)^2}\cdot\left(\sum_jd_jg^+_{ij}\vy_j^*\right) + \frac{d_i}{\text{vol}(G)}g^+_{ik}\vy_k^*\right) \nonumber\\
&=\left(g^-_{ik} - \frac{bd_i}{\text{vo}(G)}g^+_{ik}\right)\vy_k^* - \frac{b(\text{vol}(G)-d_i)}{\text{vol}(G)^2}\cdot\sum_jd_jg^+_{ij}\vy_j^*
\end{align*}
where $\frac{\partial d_i}{\partial a_{ik}} = 1$, $\frac{\partial \text{vol}(G)}{\partial a_{ik}} = 1$, $\frac{\partial d_j}{\partial a_{ik}} = \left\{
    \begin{array}{l}
      1, \text{for}\ j=k\\
      0, \text{otherwise}
    \end{array}
  \right.$.

More General, the loss function based on term $\frac{\#(w,c)}{\#w \frac{\#c}{|C|}}$ is applicable for all skip gram with negative sampling (SGNS) model based methods (LINE, PTE, DeepWalk, node2vec) \citep{levy2014neural, qiu2018network}. Based on different sampling strategy, $\frac{\#(w,c)}{\#w \frac{\#c}{|C|}}$ becomes different functions of $\ha_ij$. For example, in LINE this is ratio $\frac{\sum_i a_{ij} \cdot \sum_j a_{ij}}{\sum_{ij}a_{ij}}$. PTE is similar to but extended to multiple networks. In DeepWalk and node2vec, this expression converges probabilistically to entries of function of random walk matrix $\mP = \mD^{-1}\mA$. So in theory, the idea of ExplaiNE can be extended to all SGNS based methods.

\section{Quality of the ExplaiNE approximation}
In this section we describe the detailed evaluation about the quality of the ExplaiNE approximation. The results are obtained on GoT network with embedding dimensionality $2$.
\subsection{Compare the gradient of incident nodes v.s. non-incident nodes}
\begin{figure}[h]
  \centering
  \includegraphics[width=0.5\textwidth]{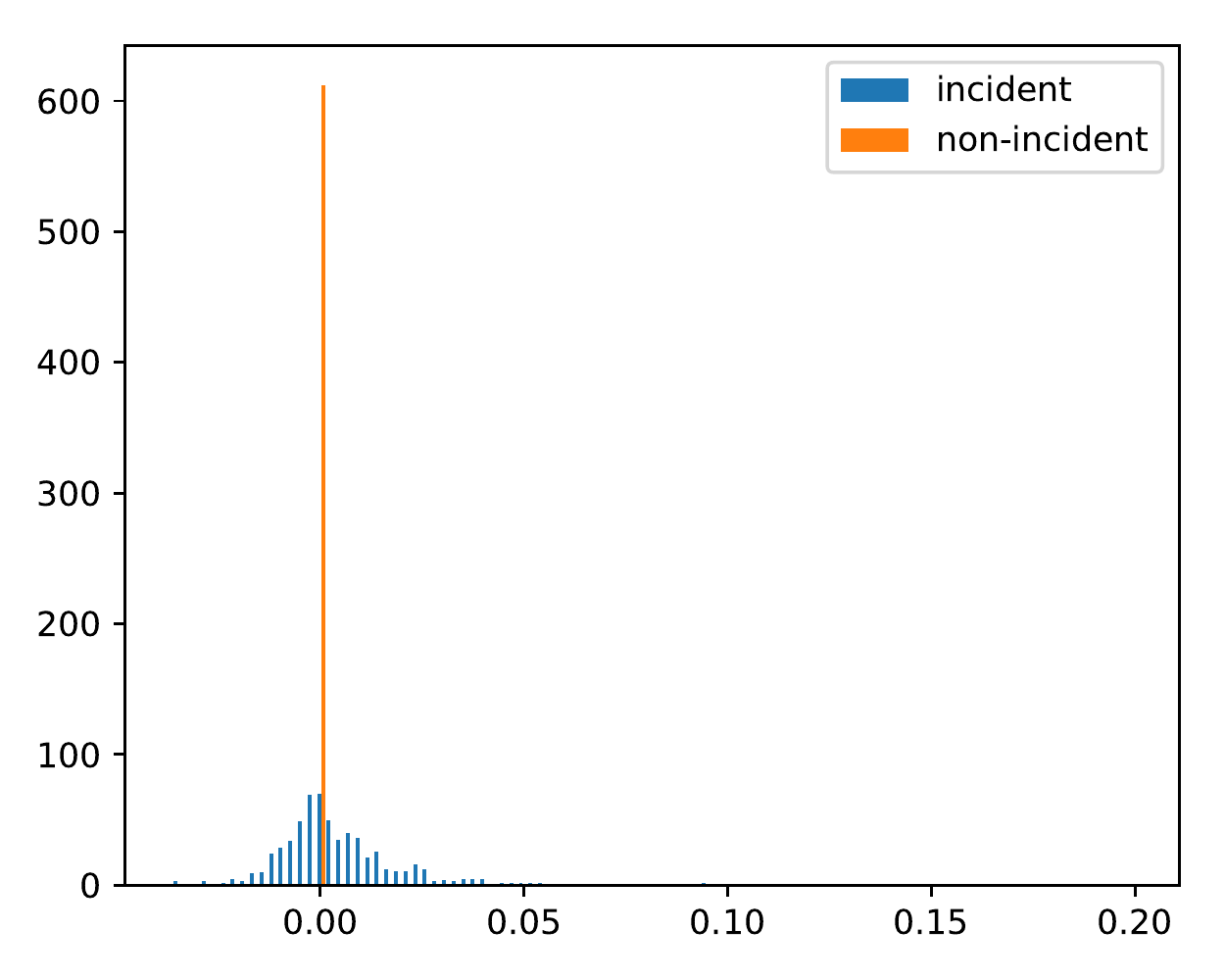}
\caption{Gradient values of (top) predicted link probability of $100$ randomly sampled nodes with respect to incident nodes (blue bars) versus non-incident nodes (orange bars). The gradient values of non-incident node pairs (on average $-2.13\mathrm{e}{-7}\pm6.96\mathrm{e}{-5}$)are mush smaller than the gradient values of the incident node pairs (on average $0.005\pm 0.02$). This validates our assumption that the derivative of the predicted link probability is higher w.r.t incident node pairs $\{i,k\}$ than w.r.t non-incident node pairs $\{k,l\}$..\label{fig:exp_grad_incident}}
\end{figure}
Our first goal is to validate the assumption (main paper Sec.2.3) that the derivative of the predicted link probability is higher w.r.t incident node pairs $\{i,k\}$ than w.r.t non-incident node pairs $\{k,l\}$. We randomly sampled $100$ nodes, for each node $i$ we compute (using exact ExplaiNE) the gradient  of the link probability of $i$'s top predicted link $\{i,j\}$ with respect to $i's$ incident pairs $\{i,k\}$. These values summarized in the histogram (Fig.\ref{fig:exp_grad_incident}) with blue bars. As comparison, we also compute (using exact ExplaiNE) for each node $i$ the gradient of predicted probability of link $\{i,j\}$ with respect to a random sample (same number as $i$'s incident nodes) of non-incident pairs $\{k,l\}$ ($k\neq l$ and $\{k,l\}\neq \{i,j\}$). These values are also summarized in the histogram (Fig.\ref{fig:exp_grad_incident}) with orange bars. The plot shows the gradient values of non-incident node pairs (on average $-2.13\mathrm{e}{-7}\pm6.96\mathrm{e}{-5}$)are mush smaller than the gradient values of the incident node pairs (on average $0.005\pm 0.02$). This validates our assumption.

\subsection{Evaluate ExplaiNE approximation on predicted links}
In this section, we assess the extent to which the top $K$ explanations for a predicted link $\{i,j\}$ incident to a given node $i$, as given by approximated ExplaiNE, overlap with the top-$K$ explanations given by exact ExplaiNE.

The relevant parameters here are (1) the value of $K$ and (2) the number of neighbors.
As we consider only links to neighbors as candidate explanations, $K$ must be smaller than the number of neighbors. Moreover, if the number of neighbors is not much larger than $K$, a substantial overlap in the top-K explanations of the exact and approximate method is not surprising. Indeed, if $i$ has $m$ neighbors, two random subset of $K$ neighbors would share $l$ elements with probability $\frac{\binom{K}{l}\binom{m-K}{K-l}}{\binom{m}{K}}$, which is large for large $l$ if $m$ is not much larger than $K$.
\begin{figure}[h]
    \centering
    \includegraphics[width=0.6\textwidth]{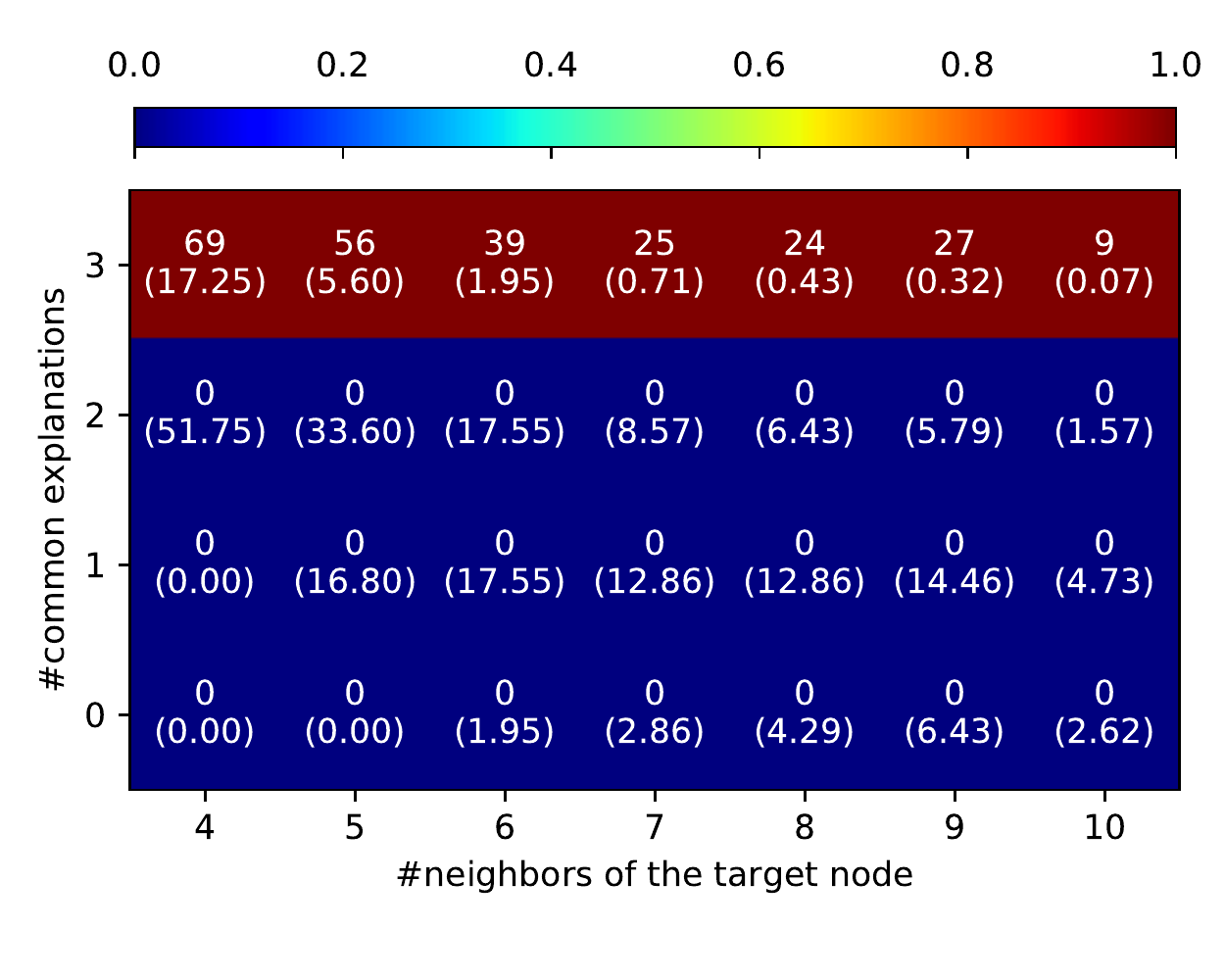}
    \caption{Histogram for $K=3$ degrees up to $10$ ($668$ out of $796$ nodes have smaller or equal degrees). It reveals that the top-$3$ explanations given by approximated ExplaiNE in almost all cases completely overlaps with the explanation given by exact ExplaiNE.\label{fig:exp_grad_overlap_0}}
\end{figure}

\begin{figure}[h]
    \centering
    \includegraphics[width=0.6\textwidth]{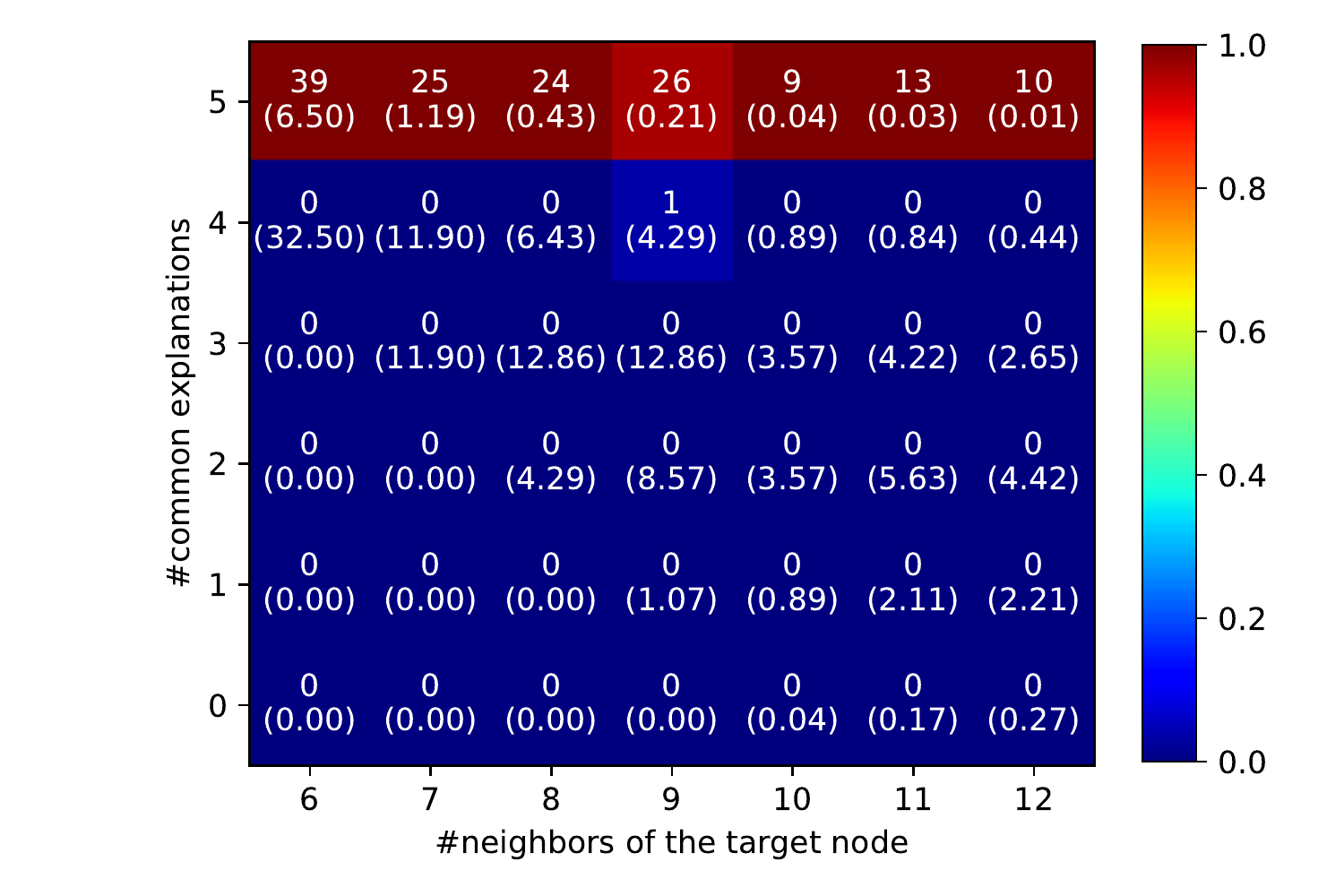}
    \caption{Histogram for $K=5$ degrees up to $12$ ($691$ out of $796$ nodes have smaller or equal degrees). It reveals that the top-$5$ explanations given by approximated ExplaiNE in almost all cases completely overlaps with the explanation given by exact ExplaiNE.\label{fig:exp_grad_overlap_1}}
\end{figure}

Thus, we performed a stratified analysis, computing the size of the overlap of the top-$K$ explanations, aggregated in a histogram over nodes with a specific degree. We plot the histogram for $K=3$ degrees up to $10$ ($668$ out of $796$ nodes have smaller or equal degrees) (Fig.~\ref{fig:exp_grad_overlap_0}) and for $K=5$ degrees up to $12$ ($691$ out of $796$ nodes have smaller or equal degrees) (Fig.~\ref{fig:exp_grad_overlap_1}). Both histogram revealed that the top-$K$ explanations given by approximated ExplaiNE in almost all cases completely overlaps with the explanation given by exact ExplaiNE.

For completeness, we counted the cases of perfect overlapping up to the largest node degree $122$ for $K = 1,\ldots,5$, summarized in Table.~\ref{tb:supp_grad_overlap}. The result agains shows the almost perfect overlapping between the explanations given by approximated ExplainNE and exact ExplainNE. Note for larger neighborhood size, the more probable the top explanations contains more noise, thus the top ranked explanations have lower similarities (e.g. in $K=5$ case) with the exact version.
\begin{table}[h]
  \caption{Number of perfect overlapping between the explanations given by approximated ExplainNE and exact ExplainNE. Count up to the largest node degree $122$ for $K = 1,\dots,5$. }
  \label{tb:supp_grad_overlap}
  \centering
  \begin{tabular}{ c c c c c c}
      & $K=1$ & $K=2$ & $K=3$ & $K=4$ & $K=5$ \\\hline
      \begin{tabular}{@{}c@{}}Number of perfect overlaps\\ out of 796 cases\end{tabular}& $796$ & $794$ & $793$ & $794$ & $789$
  \end{tabular}
\end{table}
\subsection{Evaluate ExplaiNE approximation on existing links}
\begin{figure}[h]
  \centering
  \includegraphics[width=0.6\textwidth]{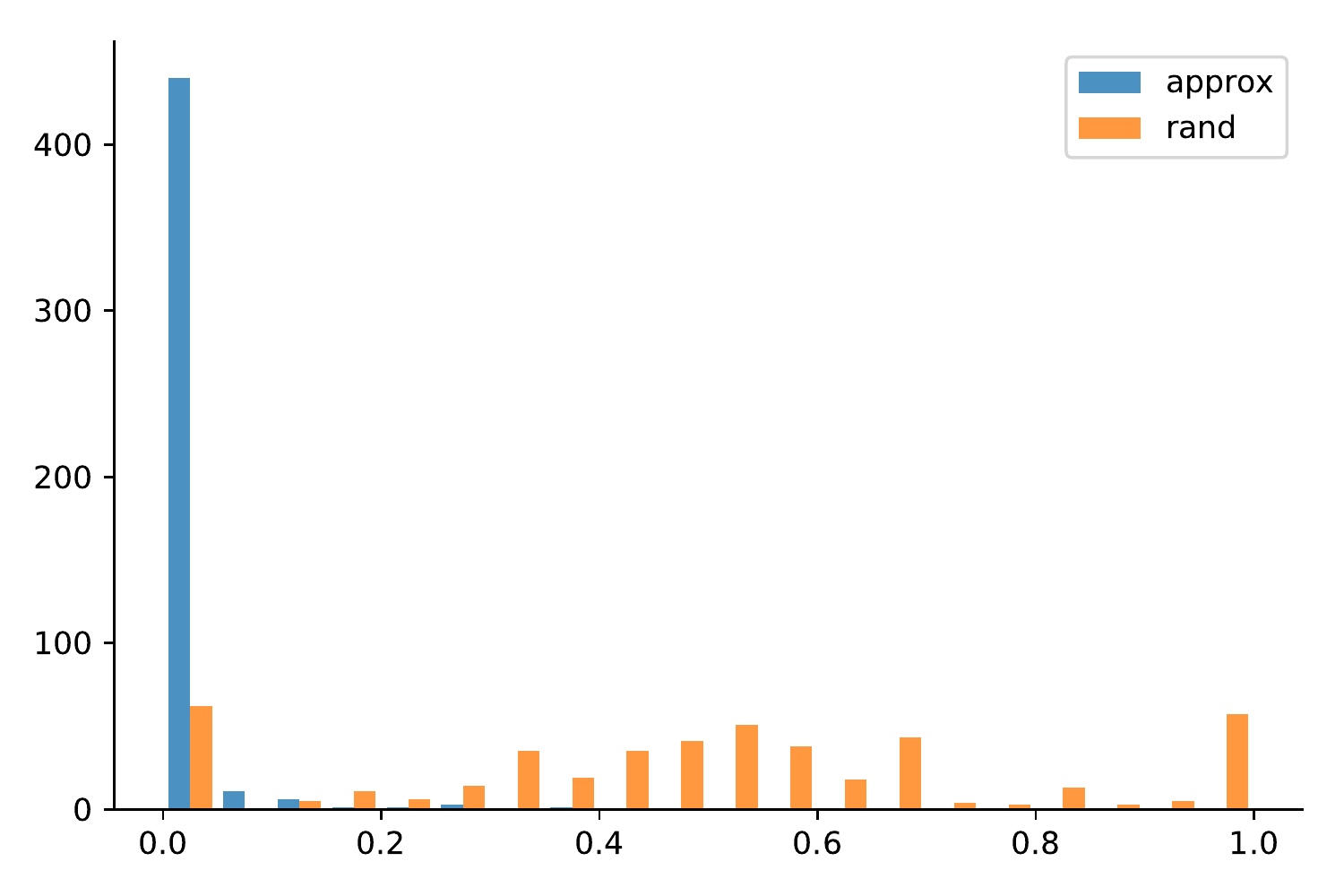}
\caption{For each node's top existing link (according to the link probability) we computed the normalized Kendall tau distance between the ranked explanation given by approximated ExplaiNE and exact ExplaiNE. We compared this with the normalized Kendall tau distance measured between a random ranking and the ranking given by exact ExplaiNE. The average normalized Kendall tau distance between explanations given by approximated and exact ExplaiNE is $0.008\pm 0.04$ and, for comparison, the average between a random ranking and exact ExplaiNE is $0.49 \pm 0.29$. This again confirms the accuracy (here, in terms of ranking distance) of approximated ExplaiNE well approximates exact version. \label{fig:supp_grad_rank}}
\end{figure}

Although the top-$K$ explanation are arguably more relevant than the tail of the ranking, for completeness we also compared the total ranking of the neighbors between the approximated and exact ExplaiNE versions, and this for the top predicted link as well as for all existing links (i.e. seeking explanations for links that are actually present in the network). More specifically, for each node's top existing link (according to the link probability) we computed the normalized Kendall tau distance between the ranked explanation given by approximated ExplaiNE and exact ExplaiNE. We compared this with the normalized Kendall tau distance measured between a random ranking and the ranking given by exact ExplaiNE (See Fig.~\ref{fig:supp_grad_rank}).The average normalized Kendall tau distance between explanations given by approximated and exact ExplaiNE is $0.008\pm 0.04$ and, for comparison, the average between a random ranking and exact ExplaiNE is $0.49 \pm 0.29$. These results again confirm the accuracy (here, in terms of ranking distance) of approximated ExplaiNE well approximates exact version.

\section{Qualitative study}
\ptitlenoskip{DBLP network} In the co-authorship network, a predicted link between authors $i$ and $j$ suggests a collaboration between them. While ExplaiNE uses no external information to provide its explanations for such suggested collaborations, our experiments indicate that such explanations tend to be existing collaborators working on a topic on which the suggested collaborator is active as well.

As first example, we predict links for ICML'19 general chair Eric P. Xing (node $i$), and compute the explanations for his top recommendation (node $j$): Adams Wei Yu. It turns out that the existing co-authors of Eric P. Xing identified by ExplaiNE as top-$5$ explanations for this recommendation (see Table~\ref{tb:supp_dblp_explanation_0}) are either colleagues or coauthors of Adams Wei Yu, with a shared interest in large scale optimization and deep learning.
\begin{table}[h]
  \caption{Predicted/recommended collaborations for Eric P. Xing. The top link (author: Adams Wei Yu) predicted by CNE are explained through co-authors of Eric P. Xing that are also colleagues or co-authors of Adams Wei Yu. The most relevant five co-authors of Eric P. Xing also cover major parts of Adams Wei Yu's research interests: large scale optimization and deep learning.}
  \label{tb:supp_dblp_explanation_0}
\centering  \begin{tabular}{ c  l l}
    \multicolumn{1}{c}{Rank} & \multicolumn{1}{c}{Recommendations} &\multicolumn{1}{c}{\begin{tabular}{@{}c@{}}Explain: \\  `Adams Wei Yu' \end{tabular}}\\\hline
    1 & Adams Wei Yu & \multicolumn{1}{c}{Hao Su} \\\cline{1-2}
    2 & Jure Leskovec & \multicolumn{1}{|c}{Li Fei-Fei} \\
    3 & Sunita Sarawagi & \multicolumn{1}{|c}{Suvrit Sra} \\
    4 & Tong Zhang & \multicolumn{1}{|c}{Fan Li} \\
    5 & Soumen Chakrabarti & \multicolumn{1}{|c}{Wei Dai} \\
  \end{tabular}
\end{table}

As second example, we predict links for ICML'19 program chair Kamalika Chaudhuri (node $i$), and compute the explanations for his top recommendation (node $j$): Matus Telgarsky. It turns out that the existing co-authors of Kamalika Chaudhuri identified by ExplaiNE as top-$5$ explanations for this recommendation (see Table~\ref{tb:supp_dblp_explanation_1}) are coauthors, advisors, or colleagues of Matus Telgarsky, with a shared interest in deep learning.
\begin{table}[h]
  \caption{Predicted/recommended collaborations for Kamalika Chaudhuri. The top link (author: Matus Telgarsky) predicted by CNE are explained through co-authors of Kamalika Chaudhuri that are also coauthors, advisors, or colleagues of Matus Telgarsky. The most relevant five co-authors of Kamalika Chaudhuri also cover major parts of Matus Telgarsky's research interests: deep learning.}
  \label{tb:supp_dblp_explanation_1}
\centering  \begin{tabular}{ c  l l}
    \multicolumn{1}{c}{Rank} & \multicolumn{1}{c}{Recommendations} &\multicolumn{1}{c}{\begin{tabular}{@{}c@{}}Explain: \\  `Matus Telgarsky' \end{tabular}}\\\hline
    1 & Matus Telgarsky & \multicolumn{1}{c}{Animashree Anandkumar} \\\cline{1-2}
    2 & Elad Hazan & \multicolumn{1}{|c}{Chicheng Zhang} \\
    3 & Majid Janzamin & \multicolumn{1}{|c}{Yoav Freund} \\
    4 & Rong Ge & \multicolumn{1}{|c}{Sanjoy Dasgupta} \\
    5 & Shai Shalev-Shwartz & \multicolumn{1}{|c}{Sham M. Kakade} \\
  \end{tabular}
\end{table}

As thrid example, we predict links for ICML'19 program chair Ruslan Salakhutdinov (node $i$), and compute the explanations for his top recommendation (node $j$): Rich Zemel. It turns out that the existing co-authors of Ruslan Salakhutdinov identified by ExplaiNE as top-$5$ explanations for this recommendation (see Table~\ref{tb:supp_dblp_explanation_2}) are students, coauthors, or colleagues of Rich Zemel, with a shared interest in deep learning.
\begin{table}[h]
  \caption{Predicted/recommended collaborations for Ruslan Salakhutdinov. The top link (author: Rich Zemel) predicted by CNE are explained through co-authors of Ruslan Salakhutdinov that are also students, coauthors, or colleagues of Rich Zemel. The most relevant five co-authors of Ruslan Salakhutdinov also cover major parts of Rich Zemel's research interests: deep learning.}
  \label{tb:supp_dblp_explanation_2}
\centering  \begin{tabular}{ c  l l}
    \multicolumn{1}{c}{Rank} & \multicolumn{1}{c}{Recommendations} &\multicolumn{1}{c}{\begin{tabular}{@{}c@{}}Explain: \\  `Rich Zemel' \end{tabular}}\\\hline
    1 & Rich Zemel & \multicolumn{1}{c}{Ryan P. Adams} \\\cline{1-2}
    2 & Ruslan Salakhudinov & \multicolumn{1}{|c}{Jimmy Ba} \\
    3 & Russell Greiner & \multicolumn{1}{|c}{Kevin Swersky} \\
    4 & Dale Schuurmans & \multicolumn{1}{|c}{Kyunghyun Cho} \\
    5 & Aaron C. Courville & \multicolumn{1}{|c}{Ryan Kiros} \\
  \end{tabular}
\end{table}
As fourth example, we predict links for Prof. Yann LeCun (node $i$), and compute the explanations for his top recommendation (node $j$): Tomas Mikolov. It turns out that the existing co-authors of Yann LeCun identified by ExplaiNE as top-$5$ explanations for this recommendation (see Table~\ref{tb:supp_dblp_explanation_3}) are either colleagues or coauthors of Tomas Mikolov, with a shared interest in deep learning.
\begin{table}[h]
  \caption{Predicted/recommended collaborations for Prof. Yann LeCun. The top link (author: Tomas Mikolov) predicted by CNE are explained through co-authors of Yann LeCun that are also students, coauthors, or colleagues of Tomas Mikolov. The most relevant five co-authors of Yann LeCun also cover major parts of Tomas Mikolov's research interests: deep learning.}
  \label{tb:supp_dblp_explanation_3}
\centering  \begin{tabular}{ c  l l}
    \multicolumn{1}{c}{Rank} & \multicolumn{1}{c}{Recommendations} &\multicolumn{1}{c}{\begin{tabular}{@{}c@{}}Explain: \\  `Tomas Mikolov' \end{tabular}}\\\hline
    1 & Tomas Mikolov & \multicolumn{1}{c}{Andrew Caplin} \\\cline{1-2}
    2 & Hans Peter Graf & \multicolumn{1}{|c}{Sumit Chopra} \\
    3 & Graham W. Taylor & \multicolumn{1}{|c}{Ido Kanter} \\
    4 & Volodymyr Mnih & \multicolumn{1}{|c}{Clement Farabet} \\
    5 & Rodolfo A. Milito & \multicolumn{1}{|c}{Wojciech Zaremba} \\
  \end{tabular}
\end{table}
As fifth example, we predict links for Prof. Michael I. Jordan (node $i$), and compute the explanations for his top recommendation (node $j$): Christos Faloutsos. It turns out that the existing co-authors of Michael I. Jordan identified by ExplaiNE as top-$5$ explanations for this recommendation (see Table~\ref{tb:supp_dblp_explanation_4}) are either colleagues or coauthors of Christos Faloutsos, with a shared interest mainly in data mining and database systems.
\begin{table}[h]
  \caption{Predicted/recommended collaborations for Prof. Michael I. Jordan. The top link (author: Christos Faloutsos) predicted by CNE are explained through co-authors of Michael I. Jordan that are also students, coauthors, or colleagues of Christos Faloutsos. The most relevant five co-authors of Michael I. Jordan also cover major parts of Christos Faloutsos's research interests: data mining and database systems}
  \label{tb:supp_dblp_explanation_4}
  \centering  \begin{tabular}{ c  l l}
      \multicolumn{1}{c}{Rank} & \multicolumn{1}{c}{Recommendations} &\multicolumn{1}{c}{\begin{tabular}{@{}c@{}}Explain: \\  `Christos Faloutsos' \end{tabular}}\\\hline
      1 & Christos Faloutsos & \multicolumn{1}{c}{Pinar Duygulu} \\\cline{1-2}
      2 & Carlos Guestrin & \multicolumn{1}{|c}{Deepayan Chakrabarti} \\
      3 & Wolfgang Maass & \multicolumn{1}{|c}{Jennifer G. Dy} \\
      4 & Andrew Zisserman & \multicolumn{1}{|c}{Richard M. Karp} \\
      5 & Kotagiri Ramamohanarao & \multicolumn{1}{|c}{Stephen Tu} \\
    \end{tabular}
\end{table}
\ptitlenoskip{MovieLens network} In the rating network, a predicted link between a user $i$ and movie $j$ amounts to a recommendation of movie $j$ to user $i$. In making this recommendation CNE did not have access to any meta-data of the users or movies, and neither does ExplaiNE to identify explanations. Yet, we can make use of this meta-data to qualitatively assess whether the explanations make sense.
As our first example, we computed the recommendation for the first user (uid=$0$) in the user list (See Table.~\ref{tb:supp_imdb_explanation_0}). The top recommended movie is `Batman' with genre tags `Action', `Adventure', `Crime', and `Drama'. The genres of the top explanations given by explainNE arguably have strongly overlapping genre tags (e.g. all top-$5$ are tagged with `Action'). Moreover, the second-highest ranked explanation is `Batman Forever'.
\begin{table}[t]
  \caption{Recommended movie to user uid=$0$. The top movie recommended by CNE (Batman) is explained through movies already seen by user uid=$0$. The top-ranked explanations have genres that overlap with the recommended movie. \label{tb:supp_imdb_explanation_0}}
\centering
  \begin{tabular}{ c  p{34mm}  p{32mm}}
    \multicolumn{1}{c}{$j$} & \multicolumn{1}{c}{Recommendations} & \multicolumn{1}{c}{Genres}\\\hline
    1 & Batman & Action, Adventure, Crime, Drama\\
    2 & E.T. the Extra-Terrestrial & Children's, Drama, Fantasy, Sci-Fi\\
    3 & The Secret of Roan Inish & Adventure \\

    \multicolumn{1}{c}{}&\multicolumn{1}{c}{}&\multicolumn{1}{c}{}\\

    $k$& \multicolumn{1}{c}{Explain: `Batman'} & \multicolumn{1}{c}{Genres}\\\hline
    1& Supercop & Action, Thriller\\
    2& Batman Forever & Action, Adventure, Comedy, Crime\\
    3& The Crow & Action, Romance, Thriller \\
    4& Full Metal Jacket & Action, Drama, War \\
    5& Young Guns & Action, Comedy, Western \\
  \end{tabular}
\end{table}
As second example, we computed the recommendation for the first user (uid=$1$) in the user list (See Table.~\ref{tb:supp_imdb_explanation_1}). The top recommended movie is `The Devil's Own' with genre tags `Action', `Drama', `Thriller',' and `War'. The genres of the top explanations given by explainNE arguably have strongly overlapping genre tags.
\begin{table}[t]
  \caption{Recommended movie to user uid=$1$. The top movie recommended by CNE (The Devil's Own) is explained through movies already seen by user uid=$1$. The top-ranked explanations have genres that overlap with the recommended movie. \label{tb:supp_imdb_explanation_1}}
\centering
  \begin{tabular}{ c  p{34mm}  p{32mm}}
    \multicolumn{1}{c}{$j$} & \multicolumn{1}{c}{Recommendations} & \multicolumn{1}{c}{Genres}\\\hline
    1 & The Devil's Own & Action, Drama, Thriller, War\\
    2 & Everyone Says I Love You & Comedy, Musical, Romance\\
    3 & Lone Star & Drama, Mystery \\

    \multicolumn{1}{c}{}&\multicolumn{1}{c}{}&\multicolumn{1}{c}{}\\

    $k$& \multicolumn{1}{c}{Explain: `The Devil's Own'} & \multicolumn{1}{c}{Genres}\\\hline
    1& Heat & Action, Crime, Thriller\\
    2& Midnight in the Garden of Good and Evil & Comedy, Crime, Drama, Mystery\\
    3& A Time to Kill & Action, Drama \\
    4& Liar Liar & Comedy \\
    5& Up Close \& Personal & Drama, Romance \\
  \end{tabular}
\end{table}

As thrid example, we computed the recommendation for the first user (uid=$1$) in the user list (See Table.~\ref{tb:supp_imdb_explanation_2}). The top recommended movie is `The Replacement Killers' with genre tags `Action' and `Thriller'. The genres of the top explanations given by explainNE arguably have descent amount of overlapping genre tags.
\begin{table}[t]
  \caption{Recommended movie to user uid=$2$. The top movie recommended by CNE (The Replacement Killers) is explained through movies already seen by user uid=$2$. The top-ranked explanations have genres that overlap with the recommended movie. \label{tb:supp_imdb_explanation_2}}
\centering
  \begin{tabular}{ c  p{34mm}  p{32mm}}
    \multicolumn{1}{c}{$j$} & \multicolumn{1}{c}{Recommendations} & \multicolumn{1}{c}{Genres}\\\hline
    1 & The Replacement Killers & Action, Thriller\\
    2 & Titanic & Action, Drama, Romance\\
    3 & The Full Monty & Comedy \\

    \multicolumn{1}{c}{}&\multicolumn{1}{c}{}&\multicolumn{1}{c}{}\\

    $k$& \multicolumn{1}{c}{Explain: `The Replacement Killers'} & \multicolumn{1}{c}{Genres}\\\hline
    1& Spice World & Comedy, Musical\\
    2& Deep Rising & Action, Horror, Sci-Fi\\
    3& Deconstructing Harry & Comedy, Drama \\
    4& Fallen & Action, Mystery, Thriller \\
    5& Wag the Dog & Comedy, Drama \\
  \end{tabular}
\end{table}
As fourth example, we computed the recommendation for the first user (uid=$1$) in the user list (See Table.~\ref{tb:supp_imdb_explanation_3}). The top recommended movie is ` Murder at 1600' with genre tags `Mystery' and `Thriller'. The genres of the top explanations given by explainNE arguably have strongly overlapping genre tags.
\begin{table}[t]
  \caption{Recommended movie to user uid=$3$. The top movie recommended by CNE (Murder at 1600) is explained through movies already seen by user uid=$3$. The top-ranked explanations have genres that overlap with the recommended movie. \label{tb:supp_imdb_explanation_3}}
\centering
  \begin{tabular}{ c  p{34mm}  p{32mm}}
    \multicolumn{1}{c}{$j$} & \multicolumn{1}{c}{Recommendations} & \multicolumn{1}{c}{Genres}\\\hline
    1 & Murder at 1600 & Mystery, Thriller\\
    2 & The Devil's Advocate & Crime, Horror, Mystery, Thriller\\
    3 & The Game & Mystery, Thriller \\

    \multicolumn{1}{c}{}&\multicolumn{1}{c}{}&\multicolumn{1}{c}{}\\

    $k$& \multicolumn{1}{c}{Explain: `Murder at 1600'} & \multicolumn{1}{c}{Genres}\\\hline
    1& Liar Liar & Comedy\\
    2& Scream & Horror, Thriller\\
    3& Cop Land & Crime, Drama, Mystery \\
    4& Assignment & Thriller \\
    5& Conspiracy Theory & Action, Mystery, Romance, Thriller \\
  \end{tabular}
\end{table}

As fifth example, we computed the recommendation for the first user (uid=$1$) in the user list (See Table.~\ref{tb:supp_imdb_explanation_4}). The top recommended movie is `Dumbo' with genre tags `Animation', `Childrens', and  `Musical'. The genres of the top explanations given by explainNE arguably have strongly overlapping genre tags (e.g. all top-$4$ are tagged with `Childrens').
\begin{table}[t]
  \caption{Recommended movie to user uid=$4$. The top movie recommended by CNE (Dumbo) is explained through movies already seen by user uid=$4$. The top-ranked explanations have genres that overlap with the recommended movie. \label{tb:supp_imdb_explanation_4}}
\centering
  \begin{tabular}{ c  p{34mm}  p{32mm}}
    \multicolumn{1}{c}{$j$} & \multicolumn{1}{c}{Recommendations} & \multicolumn{1}{c}{Genres}\\\hline
    1 & Dumbo & Animation, Children's, Musical\\
    2 & The Lion King & Animation, Children's, Musical\\
    3 & The ox and the Hound & Animation, Children's \\

    \multicolumn{1}{c}{}&\multicolumn{1}{c}{}&\multicolumn{1}{c}{}\\

    $k$& \multicolumn{1}{c}{Explain: `Dumbo'} & \multicolumn{1}{c}{Genres}\\\hline
    1& Fantasia & Animation, Children's, Musical\\
    2& Cinderella & Animation, Children's, Musical\\
    3& The Parent Trap & Children's, Drama \\
    4& Alice in Wonderland & Animation, Children's, Musical \\
    5& Jack & Comedy, Drama \\
  \end{tabular}
\end{table}
%
\section{Quantitative study}
\ptitlenoskip{DBLP network} 
Here, we can construct ground truth explanations for \emph{existing} links (as opposed to \emph{predicted} ones). While this is not the intended use case of ExplaiNE, it is perfectly legitimate and justified here given our intention to objectively validate the quality of the explanations.
Our approach is based on the intuition that a one-time co-author $j$ of a given author $i$ could have been introduced to that author $i$ by another co-author $k$ on the same paper, thus explaining the link $\{i,j\}$. While this will of course not always be true, we postulate that it is sufficiently common for ExplaiNE---providing it works well---to highlight the other co-authors as explanations for the observed link $\{i,j\}$.

Given an author $i$ and a one-time co-author $j$ of $i$, 
we used ExplaiNE to rank the other co-authors of $i$, from more to less explanatory.
We then took the top-$r$ of this ranked list as predicted co-authors on the paper $i$ co-authored with $j$. Based on this, we created a confusion matrix. Clearly, the hardness of this prediction task is different for papers with different numbers of authors. Thus, in order to get a more aggregate assessment, we summed the top-$r$ confusion matrices for all one-time co-authors of node $i$ on papers with a given number of co-authors $L$, and this for different $L$ between 3 and 5. For a given author-list length, the confusion matrices with different $r$ were then used to create precision-recall curves or ROC curves.

\begin{figure*}[t]
\centering
   \includegraphics[trim=0 0 0 0, clip=true, width=\textwidth]{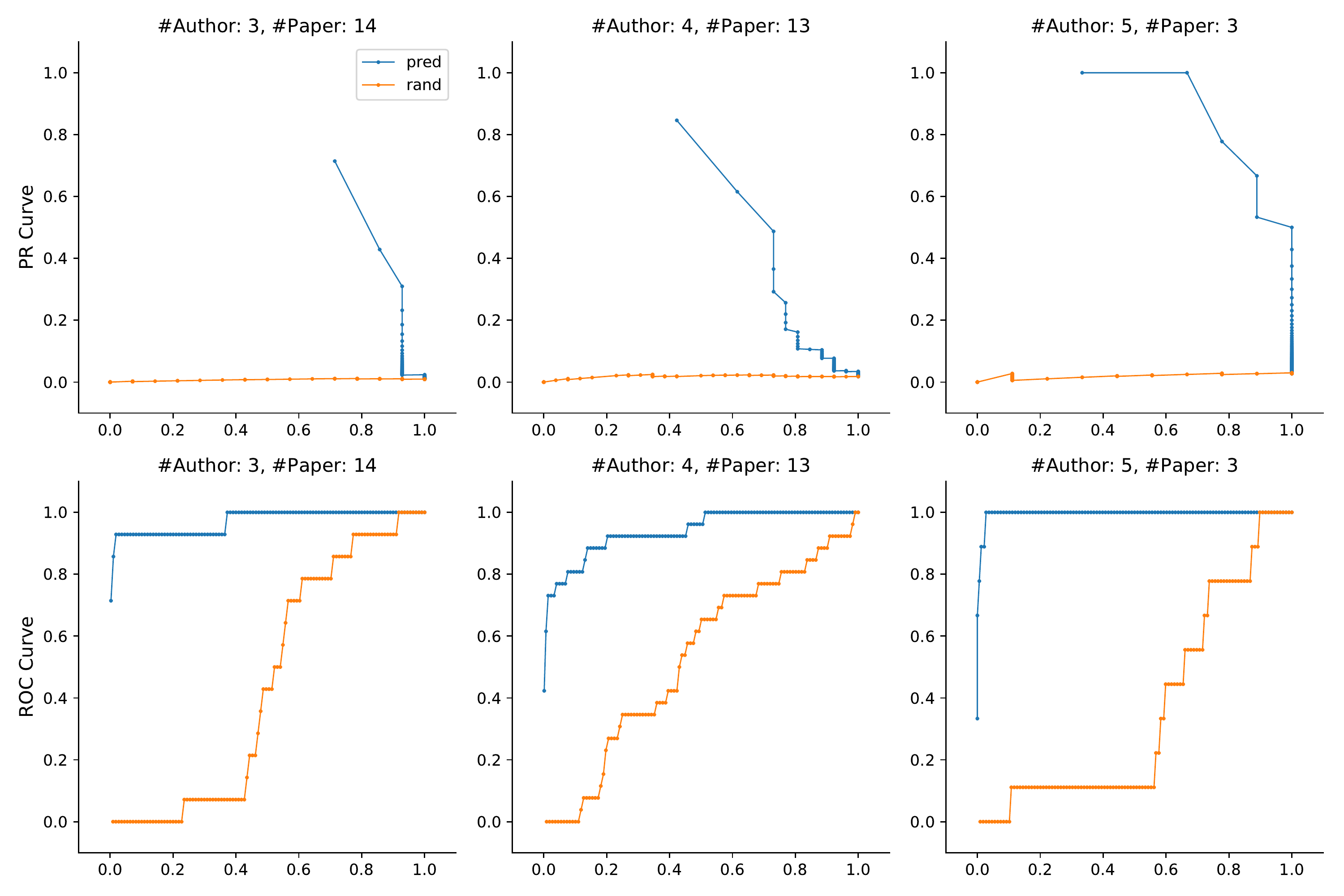}
\caption{PR (first row) and ROC (second row) curves of co-author predictions for $i=$'Eric P. Xing', with author-list lengths $3$, $4$, and $5$ (orange=random, blue=ExplaiNE). \label{fig:supp_quant_dblp_0}}
\end{figure*}

\begin{figure*}[t]
\centering
   \includegraphics[trim=0 0 0 0, clip=true, width=\textwidth]{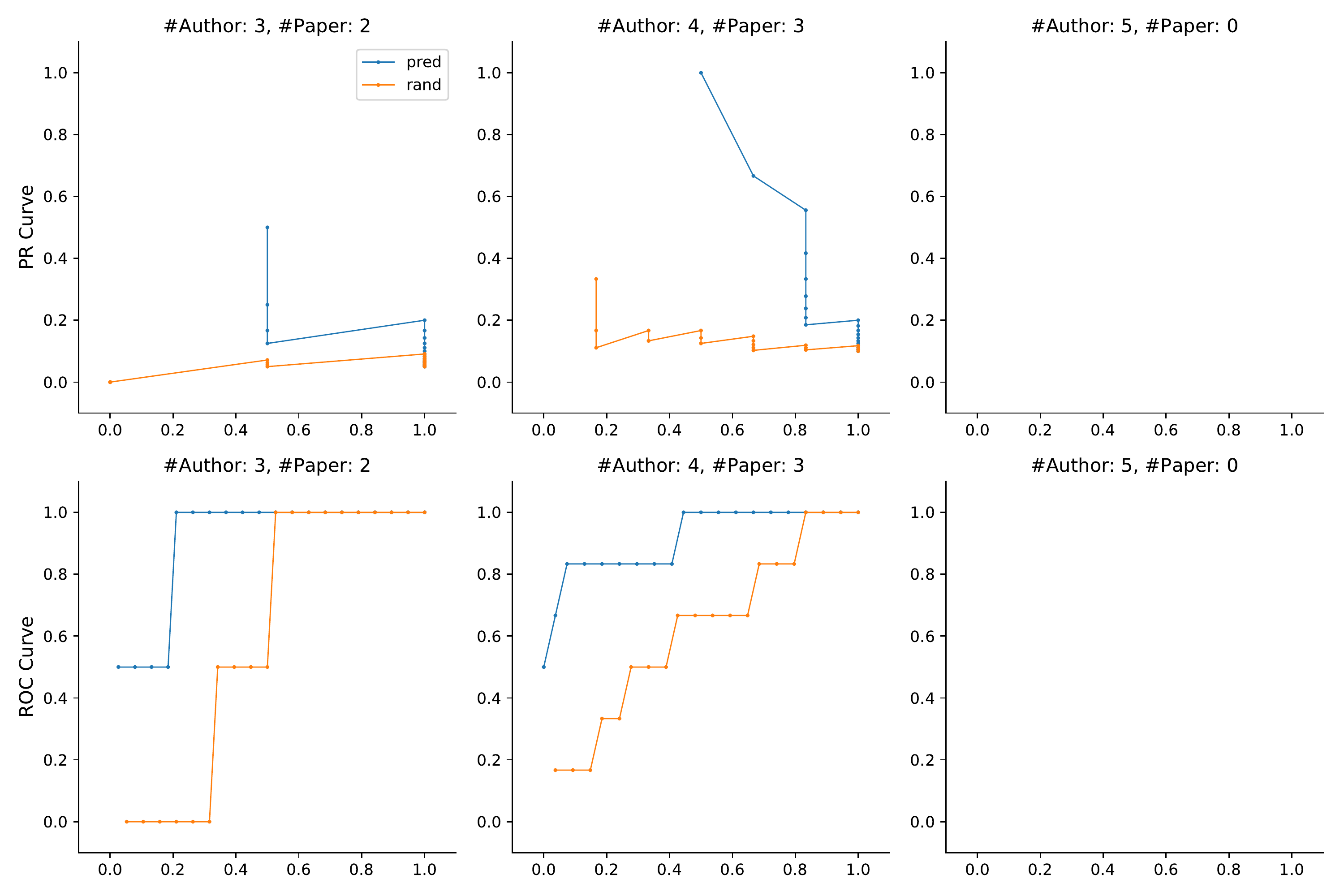}
\caption{PR (first row) and ROC (second row) curves of co-author predictions for $i=$'Kamalika Chaudhuri', with author-list lengths $3$, $4$ (orange=random, blue=ExplaiNE). In this case, no paper has author list length $5$. \label{fig:supp_quant_dblp_1}}
\end{figure*}

\begin{figure*}[t]
\centering
   \includegraphics[trim=0 0 0 0, clip=true, width=\textwidth]{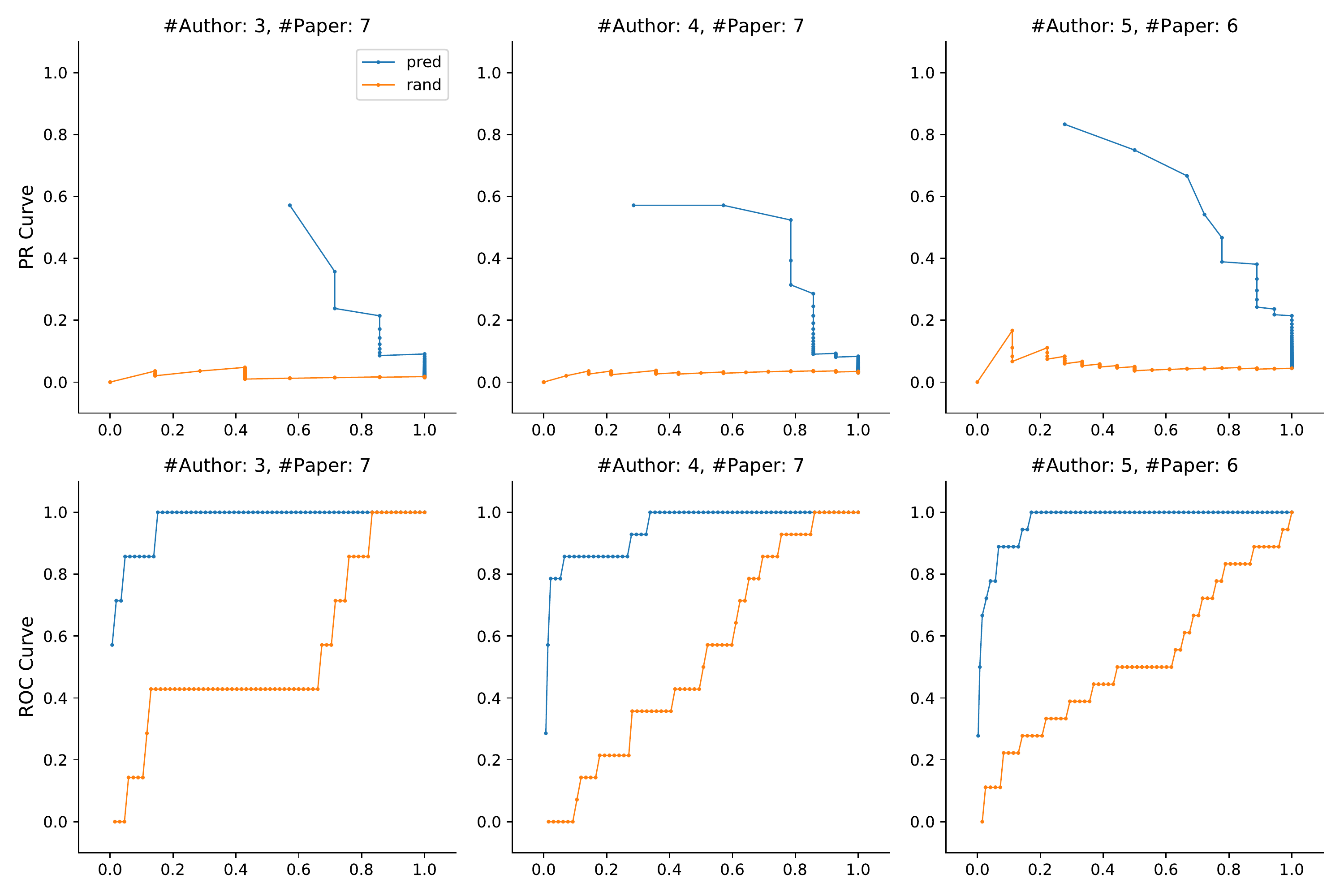}
\caption{PR (first row) and ROC (second row) curves of co-author predictions for $i=$'Ruslan Salakhutdinov', with author-list lengths $3$, $4$, and $5$ (orange=random, blue=ExplaiNE). \label{fig:supp_quant_dblp_2}}
\end{figure*}

\begin{figure*}[t]
\centering
   \includegraphics[trim=0 0 0 0, clip=true, width=\textwidth]{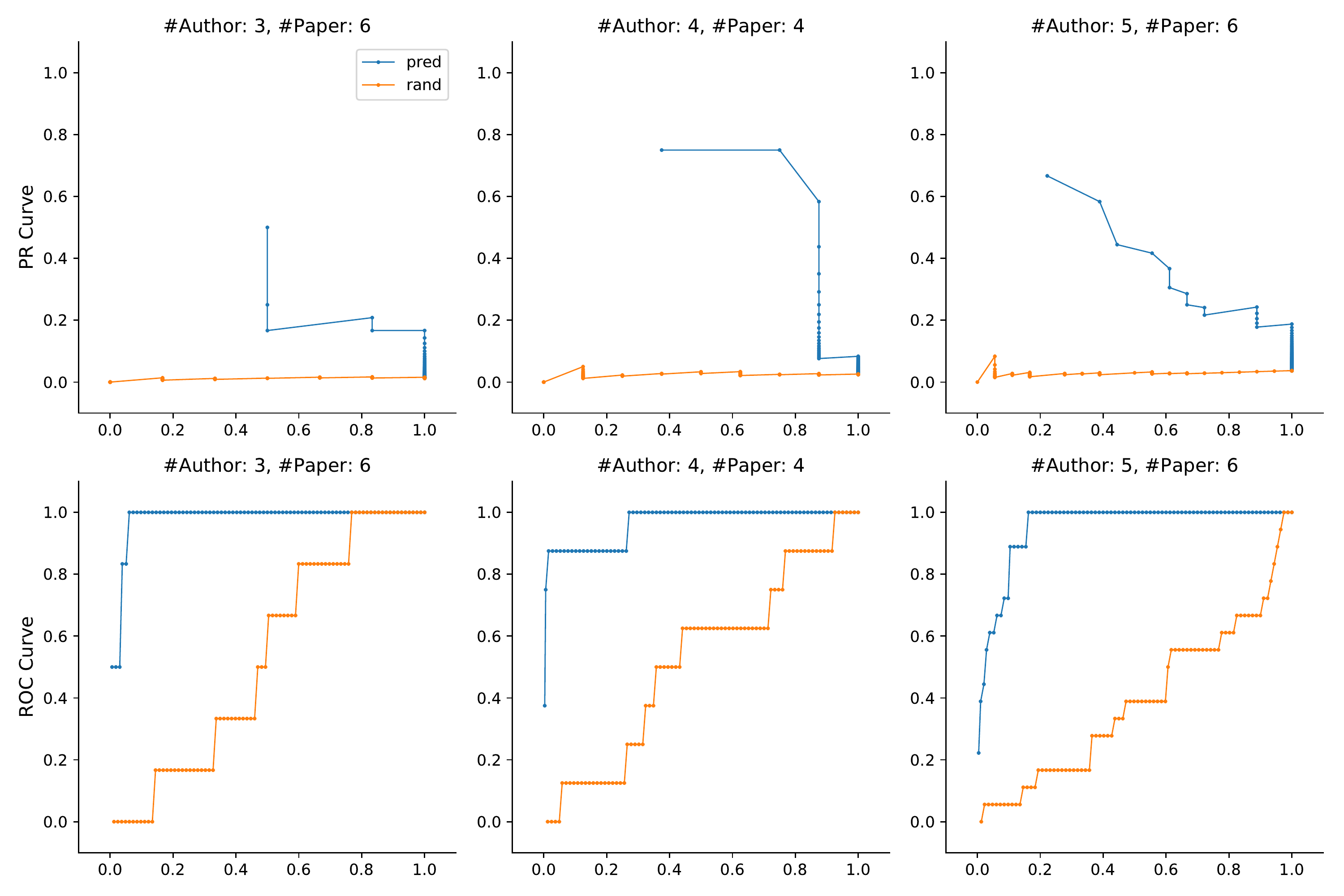}
\caption{PR (first row) and ROC (second row) curves of co-author predictions for $i=$'Yann LeCun', with author-list lengths $3$, $4$, and $5$ (orange=random, blue=ExplaiNE). \label{fig:supp_quant_dblp_3}}
\end{figure*}

\begin{figure*}[t]
\centering
   \includegraphics[trim=0 0 0 0, clip=true, width=\textwidth]{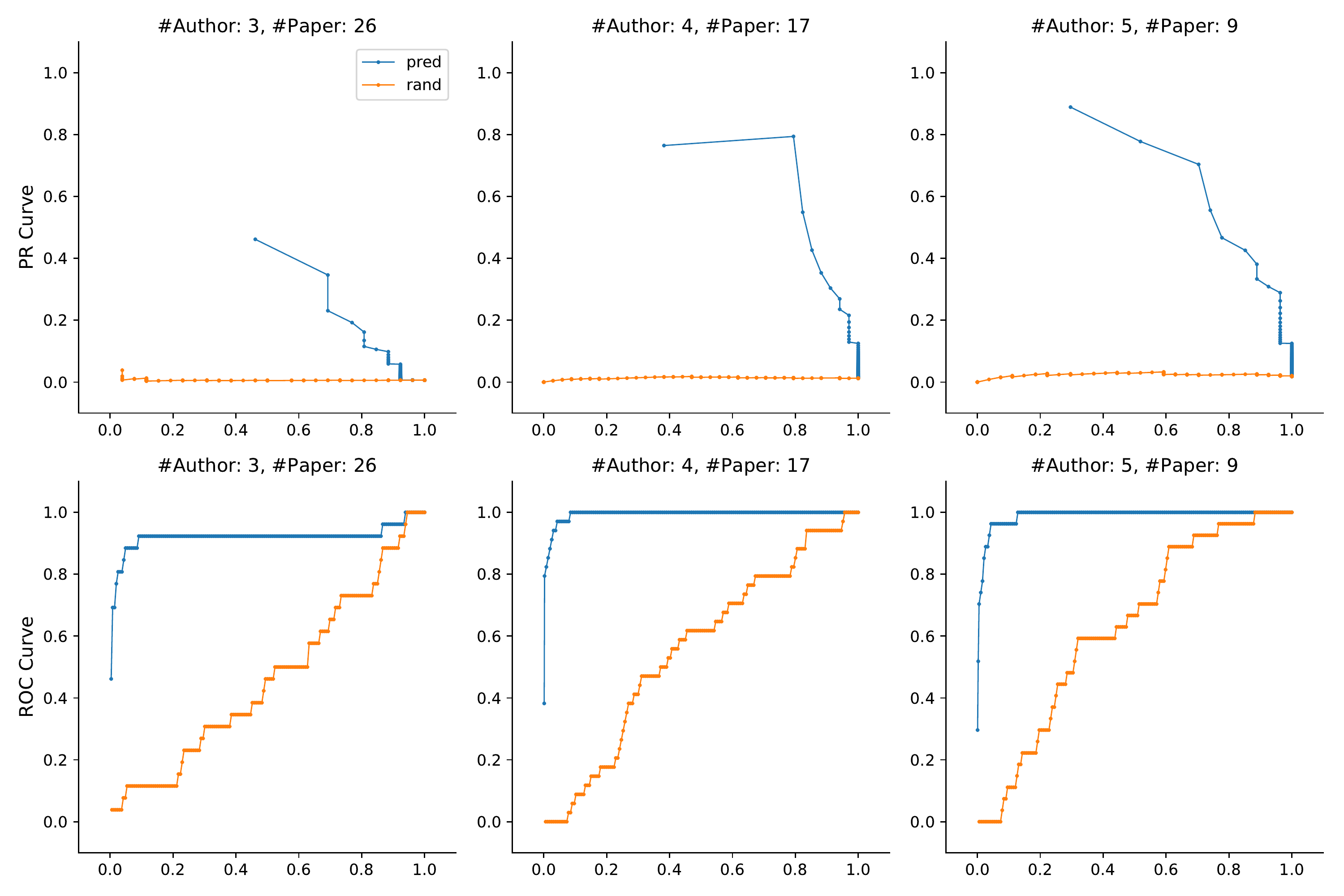}
\caption{PR (first row) and ROC (second row) curves of co-author predictions for $i=$'Michael I. Jordan', with author-list lengths $3$, $4$, and $5$ (orange=random, blue=ExplaiNE). \label{fig:supp_quant_dblp_4}}
\end{figure*}

Figure~\ref{fig:supp_quant_dblp_0} shows the PR and ROC curves for Eric P. Xing as node $i$ and three author-list lengths.
For comparison, also PR and ROC curves computed based on a randomly ranked list is shown (as the size of the data is rather small, these are not always close to the diagonal).
Figure~\ref{fig:supp_quant_dblp_1} shows the ROC curves for Kamalika Chaudhuri as node $i$ and three author-list lengths.
Figure~\ref{fig:supp_quant_dblp_2} shows the PR and ROC curves for Ruslan Salakhutdinov as node $i$ and three author-list lengths.
Figure~\ref{fig:supp_quant_dblp_3} shows the PR and ROC curves for Yann LeCun as node $i$ and three author-list lengths.
Figure~\ref{fig:supp_quant_dblp_4} shows the PR and ROC curves for Michael I. Jordan as node $i$ and three author-list lengths.

\begin{figure}[t]
  \centering
  \includegraphics[width=0.5\textwidth]{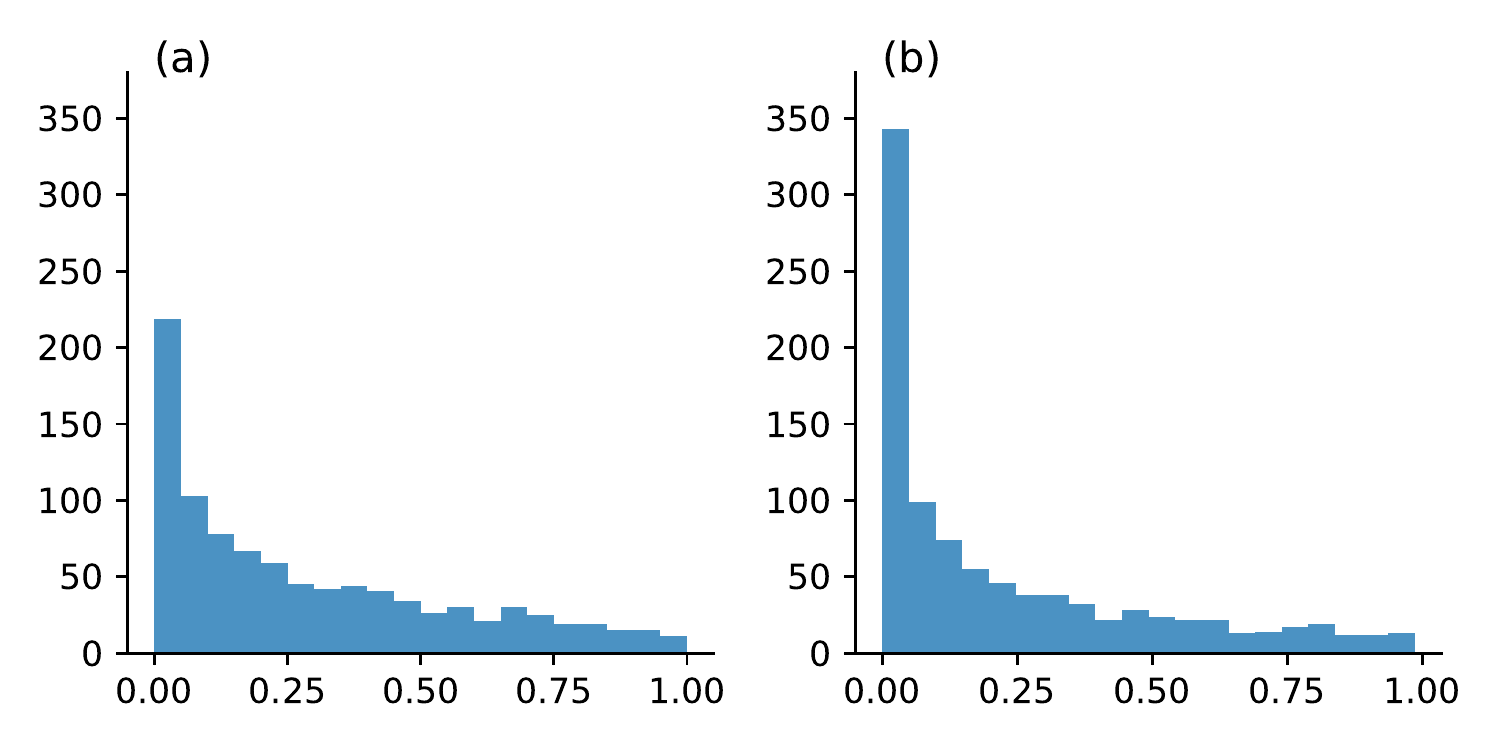}
\caption{$P$-values that indicates the significance of the correlation between the genre recommended and the genres in the explanation. Each $p$-value is computed against $50$ random explanations. (a) Explanations are drawn from user's watched movies. The empirical distribution has Kolmogorov-Smirnov test statistic $0.32$ and a $p$-value that is numerically $0.0$ against uniform distribution. (b) Explanations are drawn from all movies. The empirical distribution has Kolmogorov-Smirnov test statistic $0.42$ and a $p$-value that is numerically $0.0$ against uniform distribution. This shows the significance of positive correlation between the recommended movies and the explanations made by ExplaiNE. \label{fig:supp_quant_imdb}}
\end{figure}

\ptitlenoskip{MovieLens network} A good explanation $k$ of a predicted link between a movie-user pair $\{i,j\}$ should arguably have a similar list of genres as $j$. To test this, we computed the top-$5$ explanations for user $i$ and her top recommended movie $j$. Then we averaged the Jaccard similarity between the set of genres for movie $j$ and the set of genres of each of the $5$ explanations. To assess the significance of this average, we computed an empirical $p$-value for it by randomly sampling $50$ sets of $5$ `explanations' drawn from the watched movies of $i$, resulting in $50$ random average Jaccard similarities to compare with the one obtained by ExplaiNE. Thus we obtained an empirical $p$-value for each user $i$, indicating the significance of the overlap between the set of genres of the recommended movie $j$ and the top-$5$ explanations. We also computed the empirical $p$-values by randomly sampling $50$ sets of $5$ `explanations' drawn from the all movies for each $i$.
A histogram of these $p$-values is shown in Fig.~\ref{fig:supp_quant_imdb}a,b. While $p$-values are uniformly distributed under the null hypothesis that the explanations have genres unrelated to those of $j$, here this is not the case---indicating the null hypothesis is false. In both settings, a Kolmogorov-Smirnoff test indeed shows an extremely high significance ($p$-value numerically $0$).



\end{document}